\documentclass[runningheads]{llncs}
\usepackage{graphicx}
\usepackage{amsmath,amssymb} 
\usepackage{color}
\usepackage{graphicx}
\usepackage{amsmath}
\usepackage{amssymb}
\usepackage{booktabs}
\usepackage{tabularx}
\usepackage{multirow}
\usepackage[width=122mm,left=12mm,paperwidth=146mm,height=193mm,top=12mm,paperheight=217mm]{geometry}
\graphicspath{{./images/}{./images/plots/}}
\DeclareGraphicsExtensions{.pdf}

\usepackage[pagebackref=true,breaklinks=true,letterpaper=true,colorlinks,bookmarks=false]{hyperref}
\usepackage[dvipsnames]{xcolor}
\newcolumntype{Y}{>{\raggedright\arraybackslash}X} 

\newcommand{\myparagraph}[1]{\noindent {\bf #1.}}

\newcommand{\rtext}[2]{\parbox[t]{2mm}{\multirow{#2}{*}{\rotatebox[origin=c]{90}{#1}}}}

\def\etal{\emph{et al.}~}
\def\eg{\emph{e.g.}~}
\def\ie{\emph{i.e.}~}

\begin{document}
\pagestyle{headings}
\mainmatter

\title{Deep Image Retrieval:\\Learning global representations for image search}

\titlerunning{Learning global representations for image search}

\authorrunning{A. Gordo \and J. Almaz\'an \and J. Revaud \and D. Larlus}

\author{Albert Gordo \and Jon Almaz\'an \and Jerome Revaud \and Diane Larlus}
\institute{Computer Vision Group, Xerox Research Center Europe
  \email{firstname.lastname@xrce.xerox.com}
}

\maketitle
\begin{abstract}
We propose a novel approach for instance-level image retrieval. It produces a global and compact fixed-length
representation for each image by aggregating many region-wise descriptors. In contrast to previous works employing
pre-trained deep networks as a black box to produce features, our method leverages a deep architecture trained
for the specific task of image retrieval. Our contribution is twofold: (i)~we 
leverage a ranking framework to learn convolution and projection weights that are used to build the region features;
and (ii)~we employ a region
proposal network to learn which regions should be pooled to form the  final global descriptor. We show that
using clean training data is key to the success of our approach. To that aim, we use a large scale but noisy
landmark dataset and develop an automatic cleaning approach.  The proposed architecture produces a
global image representation in a single forward pass. Our approach significantly outperforms previous approaches based
on global descriptors on standard datasets. It even surpasses most prior works based on costly local descriptor
indexing and spatial verification\footnote{Additional material available at \texttt{www.xrce.xerox.com/Deep-Image-Retrieval}}.
\end{abstract}

\section{Introduction}
Since their ground-breaking results on image classification in recent ImageNet
challenges~\cite{Krizhevsky2012,ILSVRC15}, deep learning based methods have shined in many other
computer vision tasks, including object detection~\cite{Girshick2014} and semantic segmentation \cite{Long2015}. 
Recently, they also rekindled highly semantic tasks such as image captioning \cite{frome13devise,karpathy14deep} and visual question answering~\cite{Antol2015VQA}.
However, for some problems such as \emph{instance-level image retrieval}, deep learning methods have led to rather
underwhelming results. In fact, for most image retrieval benchmarks, the state of the art is currently held 
by conventional methods relying on local descriptor matching and re-ranking with elaborate spatial verification~\cite{Mikulik2010,Tolias2015,Tolias2015b,Xinchao2015}.

Recent works leveraging deep architectures for image retrieval are mostly limited to using a pre-trained network as local feature extractor. Most efforts have been devoted towards designing image representations suitable for image retrieval on top of those features.
This is challenging because representations for retrieval need to be compact while retaining most of the fine details of the images.
Contributions have been made to allow deep architectures to accurately represent input images of different sizes
and aspect ratios~\cite{Babenko2015,Kalantidis2015,Tolias2016} or to address the lack of geometric invariance of
convolutional neural network (CNN) features~\cite{Gong2014,Razavian2014}.

In this paper, we focus on \emph{learning} these representations. We argue that one of the main reasons for
the deep methods lagging behind the state of the art is the lack of supervised learning for the specific task of instance-level image retrieval.
At the core of their architecture, CNN-based retrieval methods often use local features extracted using networks
pre-trained on ImageNet for a classification task. These features are learned to distinguish between different semantic
categories, but, as a side effect, are quite robust to intra-class variability.  This is an undesirable
property for instance retrieval, where we are interested in distinguishing between particular objects -- even if
they belong to the same semantic category.  Therefore, learning features for the specific task of instance-level retrieval seems of paramount importance to achieve competitive results.

\begin{figure}[t!]
\includegraphics[width=1\linewidth]{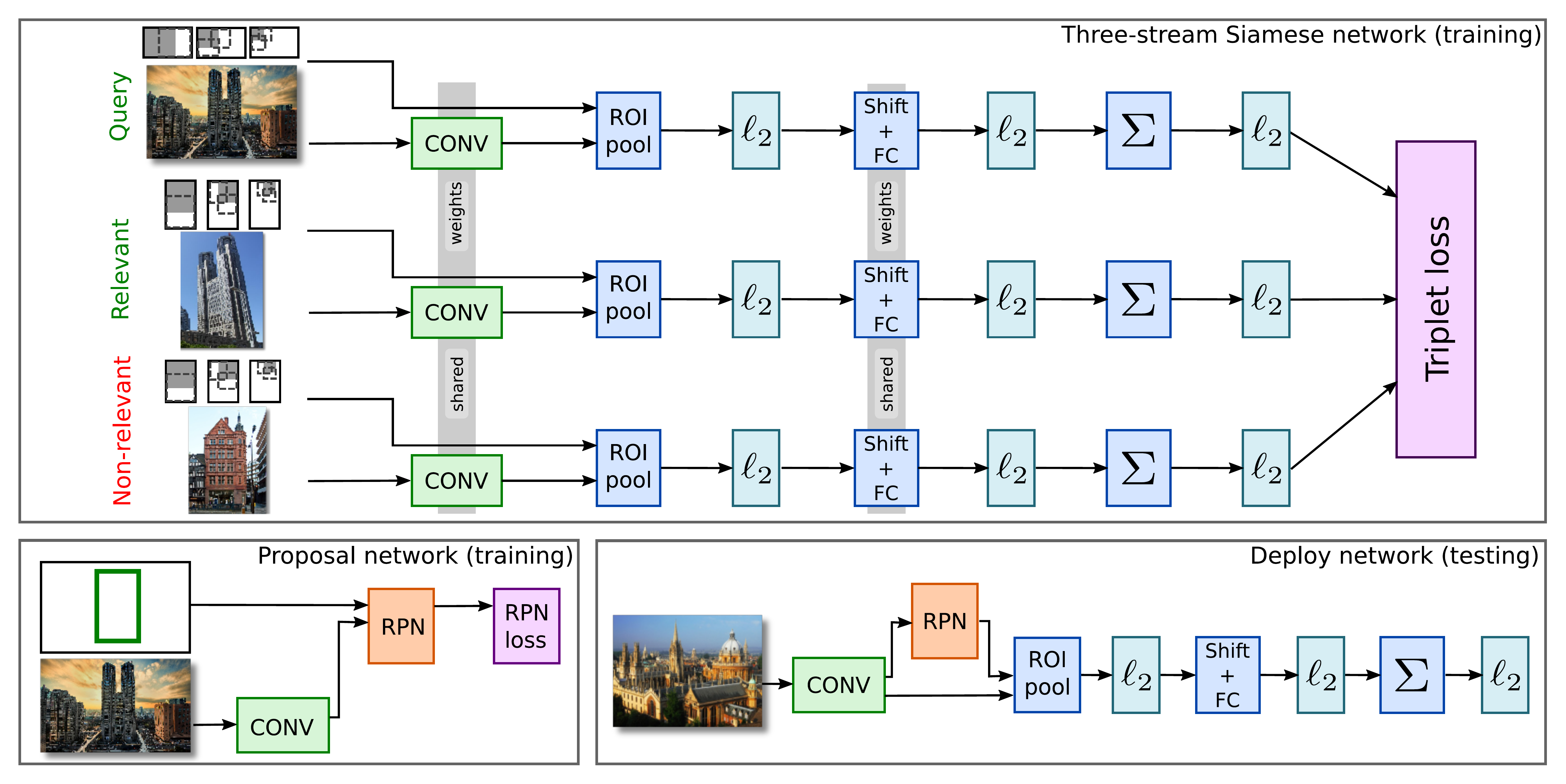}
\caption{\textbf{Summary of the proposed CNN-based representation tailored for retrieval.} At training time, image triplets are
sampled and simultaneously considered by a \textit{triplet-loss} that is well-suited for the task (top). A \textit{region proposal network} (RPN) learns which image
regions should be pooled (bottom left). At test time (bottom right), the query image is fed to the learned architecture to
efficiently produce a \textit{compact global image representation} that can be compared with the dataset image representations with a
simple dot-product.  \label{fig:method}}
\end{figure}

To this end, we build upon a recent deep representation for retrieval, the regional maximum activations of convolutions (R-MAC)~\cite{Tolias2016}. 
It aggregates several image regions into a compact feature vector of fixed length and is thus robust to scale and translation.
This representation can deal with high resolution images of different aspect ratios and obtains a competitive accuracy.
We note that all the steps involved to build the R-MAC representation are differentiable, and so its weights can be learned in an end-to-end manner.
Our \textbf{first contribution} is thus to use a \textit{three-stream Siamese network} that explicitly optimizes the weights of the R-MAC representation for the image retrieval task by using a triplet ranking loss (Fig.~\ref{fig:method}).

To train this network, we leverage the public Landmarks dataset~\cite{Babenko2014}. 
This dataset was constructed by querying image search engines with names of different landmarks and, as such, exhibits a very large amount of mislabeled and false positive images. This prevents the network from learning a good representation. 
We propose an automatic cleaning process, and show that on the cleaned data learning significantly improves.

Our \textbf{second contribution} consists in learning the pooling mechanism of the R-MAC descriptor.
In the original architecture of~\cite{Tolias2016}, a rigid grid determines the location of regions that are pooled together.
Here we propose to predict the location of these regions given the image content. 
We train a region proposal network with bounding boxes that are estimated for the Landmarks images as a by-product of the cleaning process.
We show quantitative and qualitative evidence that region proposals significantly outperform the rigid grid.

The combination of our two contributions produces a novel architecture that is able to encode one image into a compact fixed-length vector in a single forward pass. Representations of different images can be then compared using the dot-product.
Our method significantly outperforms previous approaches based on global descriptors. It even outperforms more complex approaches that involve keypoint matching and spatial verification at test time.

Finally, we would like to refer the reader to the recent work of Radenovic \etal\cite{Radenovic2016}, concurrent to ours and published in these same proceedings, that also proposes to learn representations for retrieval using a Siamese network on a geometrically-verified landmark dataset.

The rest of the paper is organized as follows. Section \ref{sec:rw} discusses related works. Sections \ref{sec:method} and \ref{sec:cleaning}
present our contributions. Section \ref{sec:exp} validates them on five different datasets. Finally
Section \ref{sec:conclusions} concludes the paper. 

\section{Related Work}
\label{sec:rw}
We now describe previous works most related to our approach.

\myparagraph{Conventional image retrieval} 
Early techniques for instance-level retrieval are based on bag-of-features representations with large vocabularies 
and inverted files~\cite{Nister2006,Philbin2007}. Numerous methods to better approximate the matching of the descriptors 
have been proposed, see \eg\cite{Mikulik2013,Jegou2010}.
An advantage of these techniques is that spatial verification can be employed to re-rank a short-list of results~\cite{Philbin2007,Perdoch2009}, yielding a significant improvement despite a significant cost.
Concurrently, methods that aggregate the local image patches have been considered. Encoding techniques, such as the
Fisher Vector~\cite{Perronnin2007}, or VLAD~\cite{Jegou2010aggregating}, combined with compression
\cite{Perronnin2010,Jegou2012,Radenovic2015} produce global descriptors that scale to larger databases at the cost of reduced accuracy.
All these methods can be combined with other post-processing techniques such as query expansion~\cite{Chum2007,Chum2011,Arandjelovic2012three}.

\myparagraph{CNN-based retrieval} After their success in classification~\cite{Krizhevsky2012}, CNN features were used as
off-the-shelf features for image retrieval~\cite{Razavian2014,Babenko2014}. Although they outperform other standard
global descriptors, their performance is significantly below the state of the art.  Several improvements were proposed
to overcome their lack of robustness to scaling, cropping and image clutter. \cite{Razavian2014}~performs
region cross-matching and accumulates the maximum similarity per query region. \cite{Babenko2015} applies sum-pooling
to whitened region descriptors. \cite{Kalantidis2015} extends~\cite{Babenko2015} by allowing cross-dimensional
weighting and aggregation of neural codes.  Other approaches proposed hybrid models involving an encoding technique such
as FV~\cite{Perronnin2015} or VLAD~\cite{Gong2014,Paulin2015}, potentially learnt as well~\cite{Arandjelovic2016} as one of their components.

Tolias \etal\cite{Tolias2016} propose R-MAC, an approach that produces a global image representation by aggregating the activation features of a CNN in a fixed layout of spatial regions. The result is a fixed-length vector representation that, when combined with re-ranking and query expansion, achieves results close to the state of the art. 
Our work extends this architecture by discriminatively learning the representation parameters and by improving the region pooling mechanism.

\myparagraph{Fine-tuning for retrieval} Babenko \etal\cite{Babenko2014} showed that models pre-trained on ImageNet for object classification could be improved by fine-tuning them on an external set of Landmarks images.
In this paper we confirm that fine-tuning the pre-trained models for the retrieval task is indeed crucial, but argue that
one should use a good image representation (R-MAC) and a ranking loss instead of a classification loss as used in \cite{Babenko2014}.

\myparagraph{Localization/Region pooling}
Retrieval methods that ground their descriptors in regions typically consider random regions~\cite{Razavian2014} or a rigid grid of regions~\cite{Tolias2016}.
Some works exploit the center bias that benchmarks usually exhibit to weight their regions accordingly~\cite{Babenko2015}.
The spatial transformer network of~\cite{Jaderberg:2015} can be inserted in CNN architectures to transform input
images appropriately, including by selecting the most relevant region for the task.
In this paper, we would like to bias our descriptor towards interesting regions without paying an extra-cost or relying on a central bias. We achieve this by
using a proposal network similar in essence to the Faster R-CNN detection method~\cite{Ren2015faster}.

\myparagraph{Siamese networks and metric learning} Siamese networks have commonly been used for metric learning \cite{Song2015}, dimensionality reduction~\cite{Hadsell2006}, learning image descriptors~\cite{Serra2015}, and performing face identification~\cite{Chopra2005,Hu2014,Sun2014}.
Recently triplet networks (i.e. three stream Siamese networks) have been considered for metric learning~\cite{Hoffer2015,Wang2014} and face identification~\cite{SchroffK2015}.
However, these Siamese networks usually rely on simpler network architectures than the one we use here, which involves pooling and aggregation of several regions.

\section{Method}
\label{sec:method}
This section introduces our method for retrieving images in large collections.
We first revisit the R-MAC representation (Section \ref{sec:learning}) showing that, despite its handcrafted nature, all of its components consist of differentiable operations.
From this it follows that one can learn the weights of the R-MAC representation in an end-to-end manner.
To that aim we leverage a three-stream Siamese network with a triplet ranking loss.
We also describe how to learn the pooling mechanism using a region proposal network (RPN) instead of relying on a rigid grid (Section \ref{sec:proposal}).
Finally we depict the overall descriptor extraction process for a given image (Section~\ref{sec:globaldesc}).

\subsection{Learning to retrieve particular objects}
\label{sec:learning}
\myparagraph{R-MAC revisited} Recently, Tolias \etal\cite{Tolias2016} presented R-MAC, 
a global image representation particularly well-suited for image retrieval. 
The R-MAC extraction process is summarized in any of the three streams of the network in Fig.~\ref{fig:method} (top).
In a nutshell, the convolutional layers of a pre-trained network (\eg VGG16 \cite{Simonyan2014}) are used to extract
activation features from the images, which can be understood as local features that do not depend on the image size or its aspect ratio. 
Local features are max-pooled in different regions of the image using a multi-scale rigid grid with overlapping cells.
These pooled region features are independently $\ell_2$-normalized, whitened with PCA and $\ell_2$-normalized again.
Unlike spatial pyramids, instead of concatenating the region descriptors, they are sum-aggregated and $\ell_2$-normalized,
producing a compact vector whose size (typically $256$-$512$ dimensions) is independent of the number of regions in the
image. 
Comparing two image vectors with dot-product can then be interpreted as an approximate many-to-many region matching.

One key aspect to notice is that \emph{all these operations are differentiable}. In particular, the spatial pooling in
different regions is equivalent to the \emph{Region of Interest} (ROI) pooling \cite{He2014}, which is differentiable
\cite{Girshick2015}. The PCA projection can be implemented with a shifting and a fully connected (FC) layer, while the
gradients of the sum-aggregation of the different regions and the $\ell_2$-normalization are also easy to compute.
Therefore, one can implement a network architecture that, given an image and the precomputed coordinates of its regions (which depend only on the image size), produces the final R-MAC representation in a single forward pass. More importantly, \emph{one can backpropagate through the network architecture to learn the optimal weights of the convolutions and the projection}.

\myparagraph{Learning for particular instances}
We depart from previous works on fine-tuning networks for image retrieval that optimize classification using cross-entropy loss \cite{Babenko2014}.
Instead, we consider a ranking loss based on image triplets. It explicitly enforces that, given a query, a relevant element to the query and a non-relevant one, the relevant one is closer to the query than the other one. 
To do so, we use a three-stream Siamese network in which the weights of the streams are shared, see Fig.~\ref{fig:method} top.
Note that the number and size of the weights in the network (the convolutional filters and the shift and projection) is independent of the size of the images, and so we can feed each stream with images of different sizes and aspect ratios.

Let $I_q$ be a query image with R-MAC descriptor $q$, $I^+$ be a relevant image with descriptor $d^+$, and $I^-$ be a
non-relevant image with descriptor $d^-$.
We define the ranking triplet loss as
\begin{equation}
L(I_q,I^+,I^-) = \frac{1}{2} \max (0, m + \|q-d^+\|^2 - \|q-d^-\|^2),
\end{equation}
where $m$ is a scalar that controls the margin. 
Given a triplet with non-zero loss, the gradient is back-propagated through the three streams of the
network, and the convolutional layers together with the ``PCA'' layers -- the shifting and the fully connected layer -- get updated.

This approach offers several advantages. First and foremost, we directly optimize a ranking objective. Second, we can
train the network using images at the same (high) resolution that we use at test time%
\footnote{By contrast, fine-tuning
networks such as VGG16 for classification using high-resolution images is not straightforward.}. Last, learning the
optimal ``PCA'' can be seen as a way to perform discriminative large-margin metric learning \cite{Weinberger:2009} in
which one learns a new space where relevant images are closer.

\subsection{Beyond fixed regions: proposal pooling}
\label{sec:proposal}
The rigid grid used in R-MAC~\cite{Tolias2016} to pool regions tries to ensure that the object of interest is covered by at least one of the regions.
However, this uniform sampling poses two problems. First, as the grid is independent of the image content, it is
unlikely that any of the grid regions accurately align with the object of interest.
Second, many of the regions only cover background. This is problematic as the comparison between R-MAC signatures can be seen as a many-to-many region matching: image clutter will negatively affect the performance. Note that both problems are coupled: increasing the number of grid regions improves the coverage, but also the number of irrelevant regions.

We propose to replace the rigid grid with region proposals produced by a Region Proposal Network (RPN) 
trained to localize regions of interest in images. 
Inspired by the approach of Ren \etal\cite{Ren2015faster}, we model this process with a 
fully-convolutional network built on top of the convolutional layers of R-MAC (see bottom-left part of Fig.~\ref{fig:method}). 
This allows one to get the region proposals at almost zero cost.
By using region proposals instead of the rigid grid we address both problems.
First, the region proposals typically cover the object of interest more tightly than the rigid grid.
Second, even if they do not overlap exactly with the region of interest, most of the proposals do overlap significantly with it (see Section~\ref{sec:exp-proposal}), which means that increasing the number of proposals per image not only helps to increase the coverage but also helps in the many-to-many matching.

The main idea behind an RPN is to predict, for a set of candidate boxes of various sizes and aspects ratio, and at all possible image locations,
a score describing how likely each box contains an object of interest.
Simultaneously, for each candidate box it performs regression to improve its location.
This is achieved by a fully-convolutional network consisting of a first layer that uses $3\times 3$ filters, and two sibling convolutional layers with $1\times 1$ filters that predict, for each candidate box in the image, both the \emph{objectness} score and the regressed location. Non-maximum suppression is then performed on the ranked boxes to produce $k$ final proposals per image that are used to replace the rigid grid.

To train the RPN, we assign a binary class label to each candidate box, depending on how much
the box overlaps with the ground-truth region of interest, and we minimize an objective function with a multi-task loss that combines a
classification loss (log loss over object \emph{vs} background classes)
and a regression loss (smooth $\ell_1$ \cite{Girshick2015}). This is then optimized by backpropagation and
stochastic gradient descent (SGD). For more details about the implementation and the training procedure of the RPNs, we refer the reader to \cite{Ren2015faster}.

We note that one could, in principle, learn the RPN \emph{and} the ranking of the images simultaneously. However,
preliminary experiments showed that correctly weighting both losses was difficult and led to unstable results. 
In our experiments, we first learn the R-MAC representation using a rigid grid, and only then we fix the convolutional layers and learn the RPN, which replaces the rigid grid.

\subsection{Building a global descriptor}
\label{sec:globaldesc}
At test time, one can easily use this network to represent a high-resolution image. 
One feeds the image to the network, which produces the region proposals, pools the features inside the
regions, embeds them into a more discriminative space, aggregates them, and normalizes them. All these operations happen in a single forward pass
(see bottom-right part of Fig.~\ref{fig:method}). This process is also quite efficient: we can encode approximately 5 high-resolution (\ie $724$ pixels for the largest side) images per second using a single Nvidia K40 GPU.

\section{Leveraging large-scale noisy data}
\label{sec:cleaning}
\begin{figure}[t!]
\begin{centering}
\resizebox{\linewidth}{!}{\includegraphics[height=0.27\linewidth]{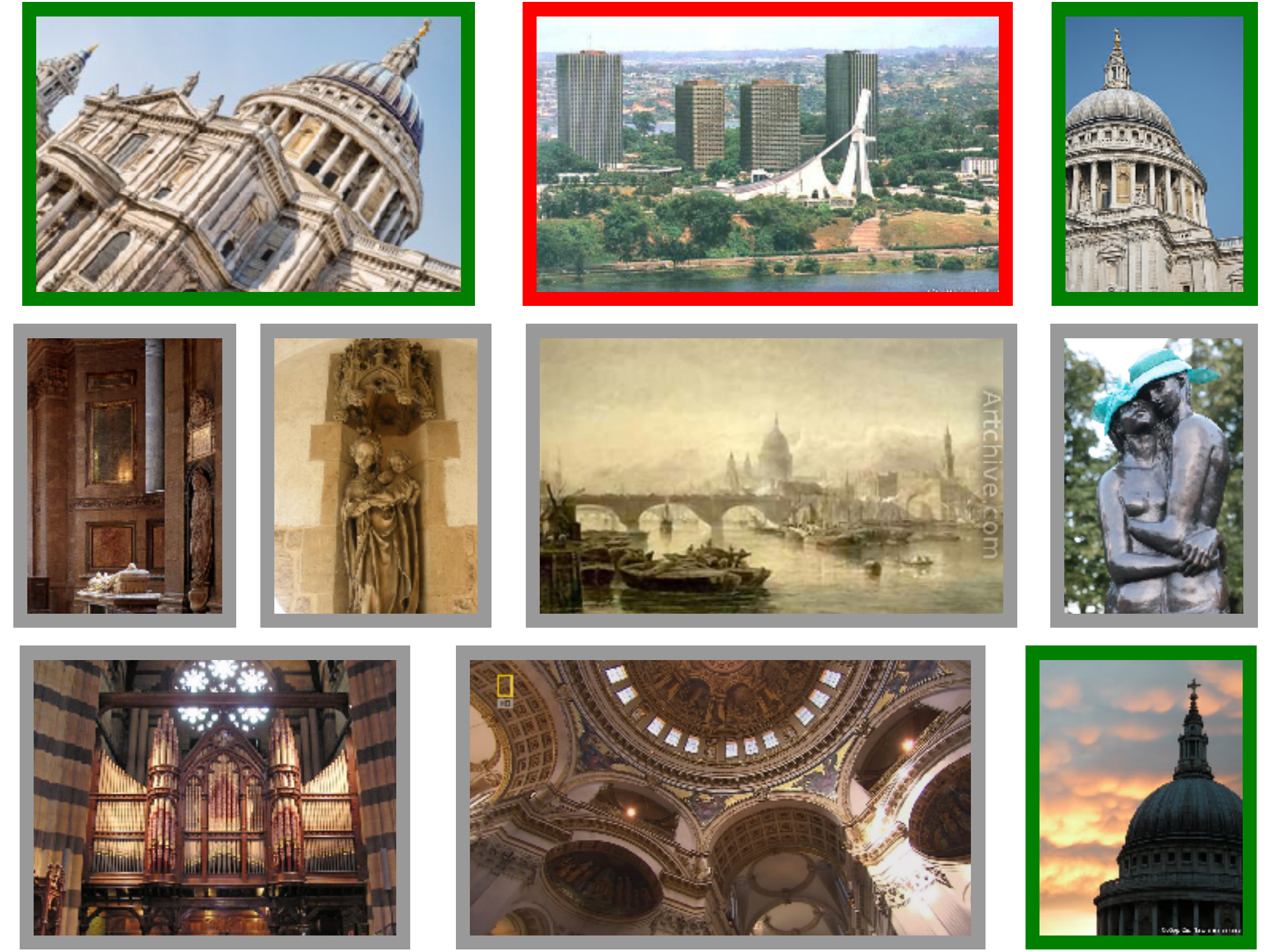}\hspace{12mm}\includegraphics[bb=30bp 150bp 930bp 530bp,clip,height=0.27\linewidth]{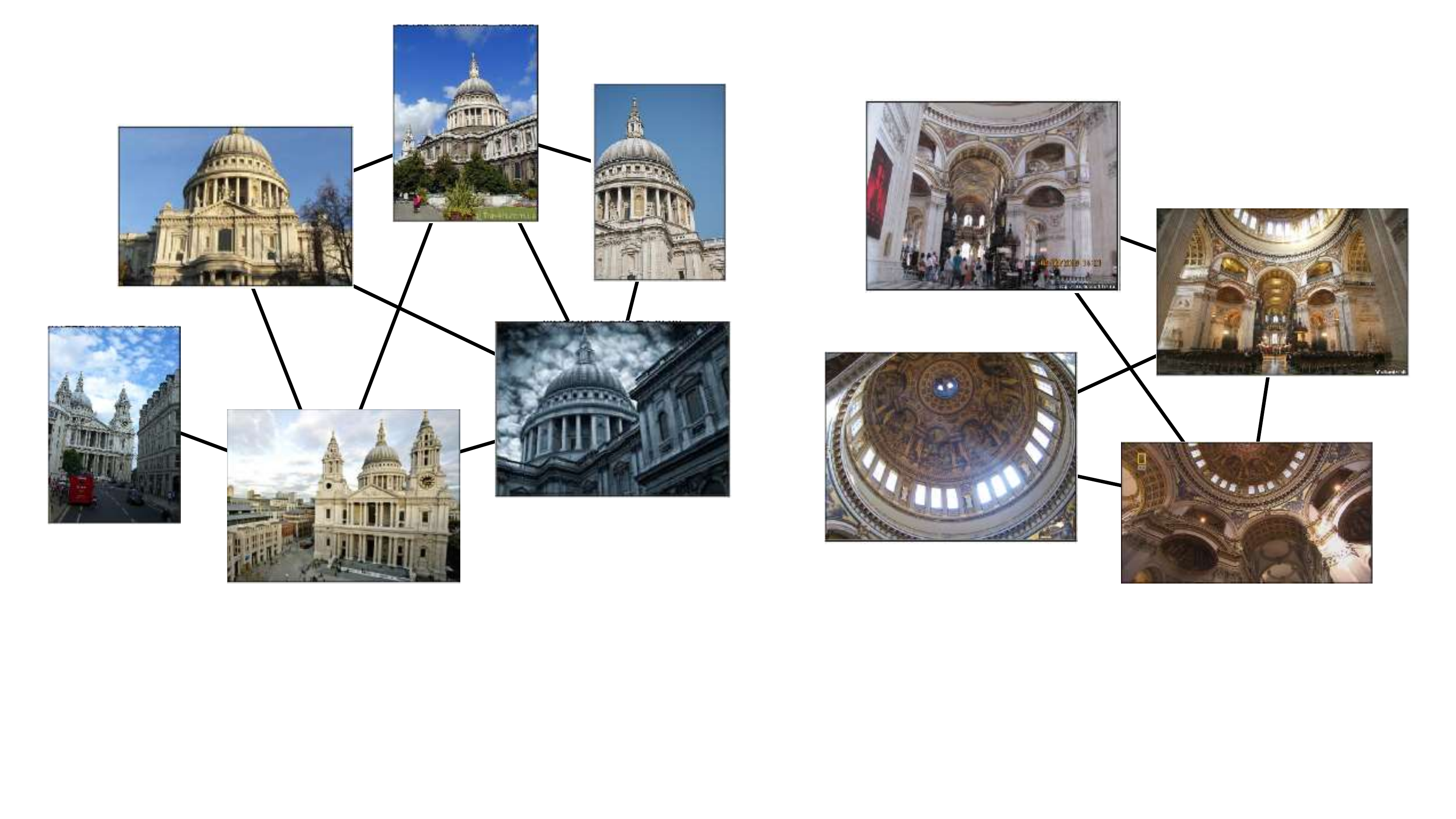}}
\par\end{centering}
\caption{\label{fig:cc}\textbf{Left}: random images from the ``St Paul's Cathedral''
landmark. Green, gray and red borders resp. denote prototypical, non-prototypical, and incorrect images.
\textbf{Right}: excerpt of the two largest connected components of the pairwise matching graph (corresponding
to outside and inside pictures of the cathedral).}
\end{figure}

To train our network for instance-level image retrieval we leverage a large-scale image dataset, the
\textbf{Landmarks} dataset~\cite{Babenko2014}, that contains approximately $214$K images 
of $672$ famous landmark sites. 
Its images were collected through textual queries in an image search engine without thorough verification.
As a consequence, they comprise a large variety of profiles:
general views of the site, close-ups of details like statues or paintings, 
with all intermediate cases as well, but also site map pictures, artistic drawings, or even completely unrelated images, see Fig.~\ref{fig:cc}.

We could only download a subset of all images due to broken URLs. After manual inspection, we merged some classes
together due to partial overlap.  We also removed classes with too few images. Finally, we meticulously removed all
classes having an overlap with the Oxford 5k, Paris 6k, and Holidays datasets, on which we experiment, see
Section~\ref{sec:exp}.
We obtained a set of about 192,000 images divided into $586$ landmarks.
We refer to this set as \textbf{Landmarks-full}.
For our experiments, we use 168,882 images for the actual fine-tuning, and the 20,668 remaining ones to validate parameters.

\myparagraph{Cleaning the Landmarks dataset} 
As we have mentioned, the Landmarks dataset present a large intra-class variability, with a wide variety of views and profiles, and
a non-negligible amount of unrelated images 
(Fig.~\ref{fig:cc}). While this is not a problem when aiming for classification (the network can accommodate during
training for this diversity and even for noise), for instance-level matching we need to train the network with images of the same
particular object or scene. In this case, variability comes from different viewing scales, angles, lighting conditions
and image clutter.
We pre-process the Landmarks dataset to achieve this as follows.

We first run a strong image
matching baseline within the images of each landmark class.  We compare each pair of images using invariant keypoint
matching and spatial verification~\cite{Lowe2004}. We use the SIFT and Hessian-Affine keypoint detectors
\cite{Lowe2004,Mikolajczyk2004} and match keypoints using the first-to-second neighbor ratio rule~\cite{Lowe2004}. This is known
to outperform approaches based on descriptor quantization~\cite{Philbin2010}. Afterwards, we verify all matches with an
affine transformation model~\cite{Philbin2007}. This heavy procedure is affordable as it is performed offline only once at training time.

Without loss of generality, we describe the rest of the cleaning procedure for a single landmark class. Once we have
obtained a set of pairwise scores between all image pairs, we construct a graph whose nodes are the images and edges are
pairwise matches.  We prune all edges which have a low score.
Then we extract the connected components of the graph.  They correspond to different
profiles of a landmark; see Fig.~\ref{fig:cc} that shows the two largest connected components for St Paul's Cathedral.
In order to avoid any confusion, we only retain the largest connected component and discard the rest.
This cleaning process leaves about 49,000 images (divided in 42,410 training and 6382 validation images) still belonging
to one of the 586 landmarks, referred to as \textbf{Landmarks-clean}.

\myparagraph{Bounding box estimation}
Our second contribution (Section~\ref{sec:proposal}) is 
to replace the uniform sampling of regions in
the R-MAC descriptor by a learned ROI selector.
This selector is trained using bounding box annotations that we automatically estimate for all landmark images.
To that aim we leverage the data obtained during the cleaning step.
The position of verified keypoint matches is a meaningful cue since the object of interest is consistently
visible across the landmark's pictures, whereas distractor backgrounds or foreground objects are varying and hence
unmatched.

\begin{figure}[t!]
\begin{centering}
\resizebox{\linewidth}{!}{\includegraphics[bb=210bp 80bp 770bp 460bp,clip,height=0.3\linewidth]{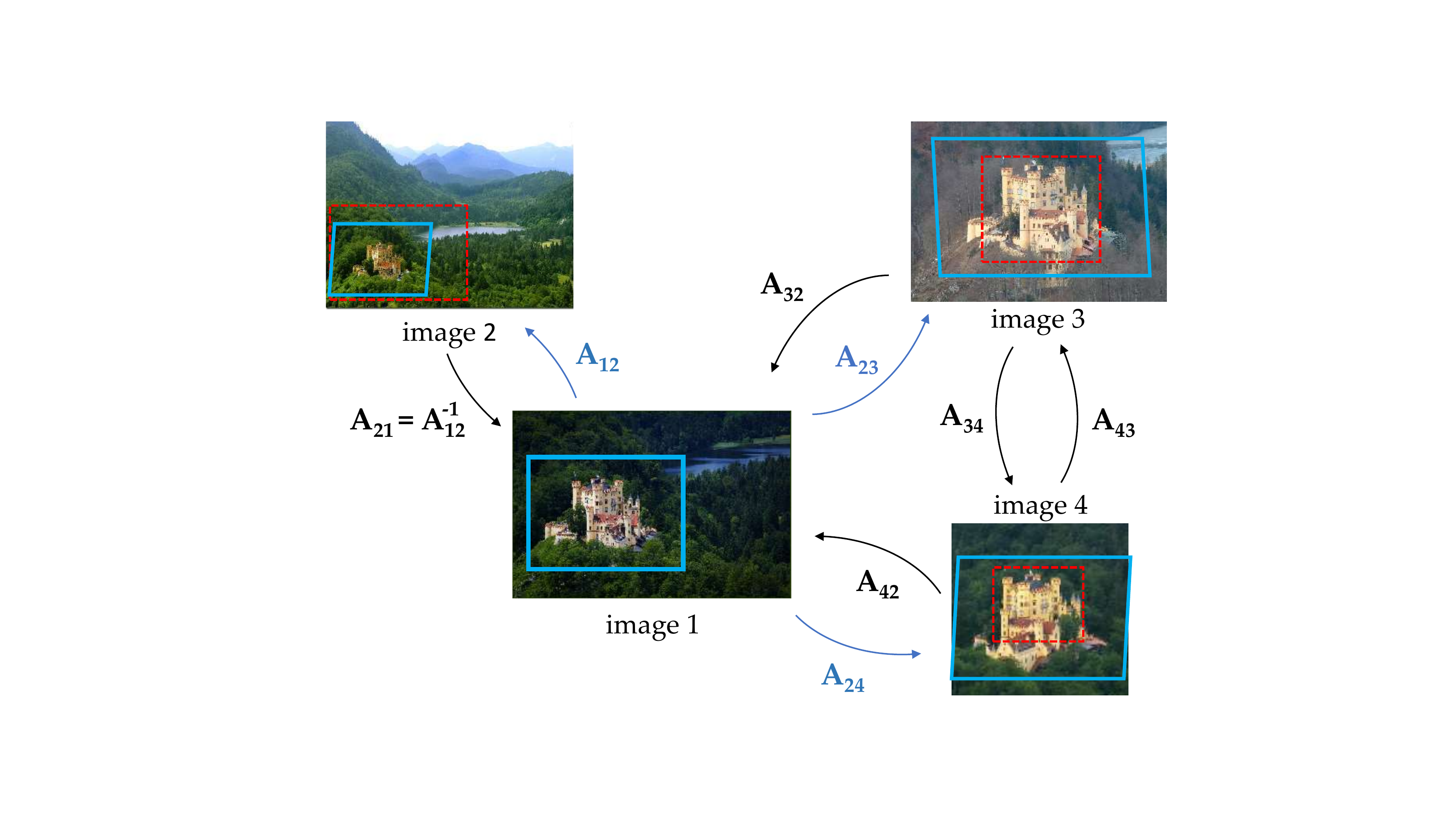}
\hspace{10mm}\includegraphics[bb=25bp 40bp 560bp 360bp,clip,height=0.3\linewidth]{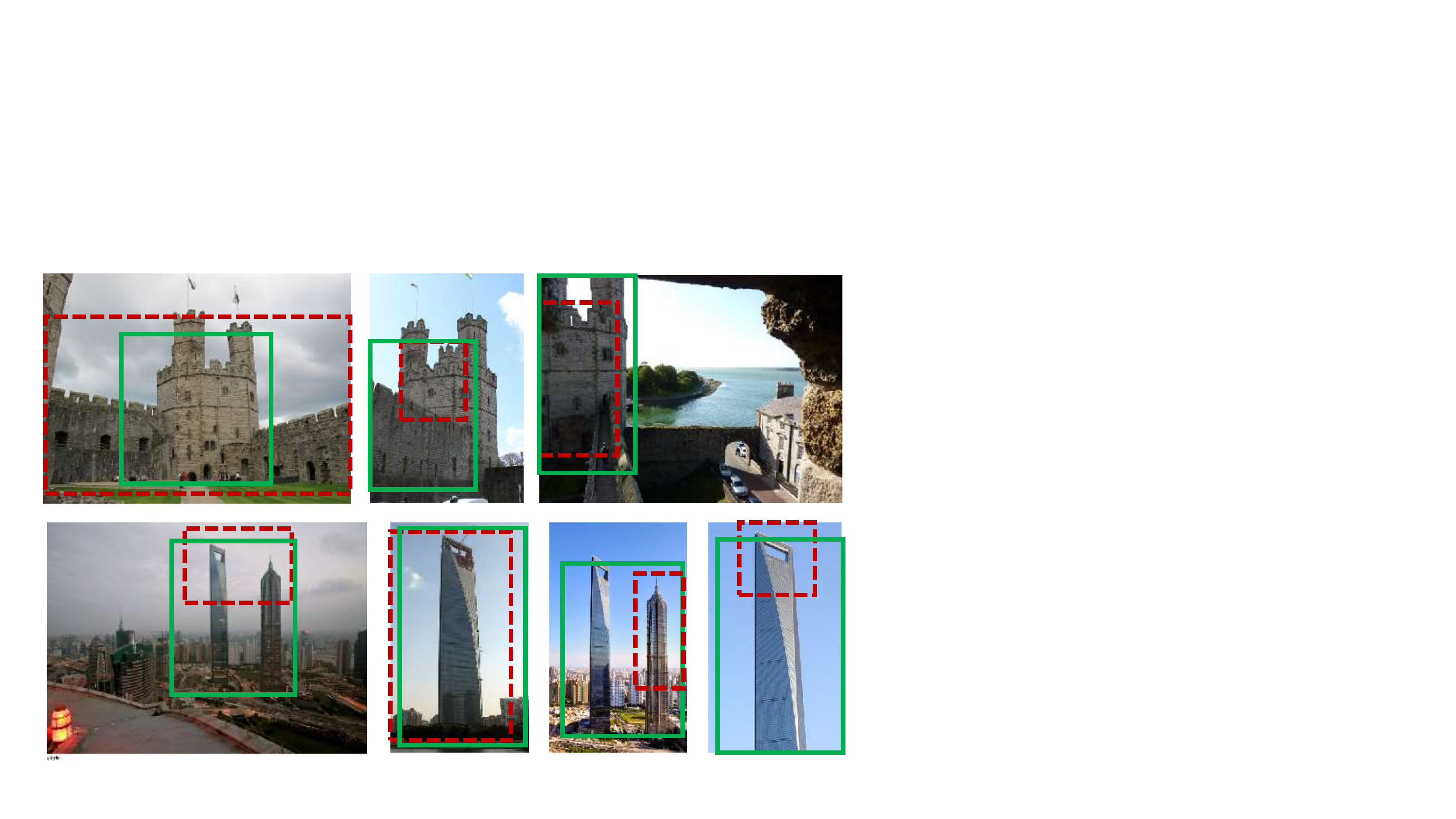}}
\par\end{centering}
\caption{\label{fig:diffusion}\textbf{Left}: the bounding box from image 1 is projected
into its graph neighbors using the affine transformations (blue rectangles). 
The current bounding box estimates (dotted red rectangles) are then updated accordingly.
The diffusion process repeats through all edges until convergence.
\textbf{Right}: initial and final bounding box estimates (resp. dotted red and plain green rectangles).}
\end{figure}

We denote the union of the connected components from all landmarks as a graph $\mathcal{S}=\left\{
\mathcal{V}_{\mathcal{S}},\mathcal{E}_{\mathcal{S}}\right\} $.  For each pair of connected images
$(i,j)\in\mathcal{E}_{\mathcal{S}}$, we have a set of verified keypoint matches with a corresponding affine
transformation $A_{ij}$.  We first define an initial bounding box in both images $i$ and $j$, denoted by $B_{i}$ and
$B_{j}$, as the minimum rectangle enclosing all matched keypoints. Note that a single image can be involved in many
different pairs. In this case, the initial bounding box is the geometric median of all boxes\footnote{Geometric median
is robust to outlier boxes compared to \emph{e.g.} averaging.}, efficiently computed
with~\cite{GeoMedian2004}. Then, we run a diffusion process, illustrated in Fig.~\ref{fig:diffusion}, in which for a
pair $(i,j)$ we predict the bounding box $B_{j}$ using $B_{i}$ and the affine transform $A_{ij}$ (and conversely).  At
each iteration, bounding boxes are updated as: $B_{j}'=(\alpha-1)B_{j}+\alpha A_{ij}B_{i}$, where $\alpha$ is a small
update step (we set $\alpha=0.1$ in our experiments). Again, the multiple updates for a single image are merged using
geometric median, which is robust against poorly estimated affine transformations.  This process iterates until
convergence. As can be seen in Fig.~\ref{fig:diffusion}, the locations of the bounding boxes are improved as well as 
their consistency across images.

\section{Experiments}
\label{sec:exp}
We now present our experimental results.
We start by describing the datasets and experimental details (Section~\ref{sec:dataset}).
We then evaluate our proposed ranking network (Section~\ref{sec:exp-ml}) and the region proposal pooling (Section~\ref{sec:exp-proposal}).
Finally, we compare our results to the state of the art  (Section~\ref{sec:exp-final}).

\begin{table}[t!]
\caption{Comparison of R-MAC \cite{Tolias2016}, our reimplementation of it and the learned versions fine-tuned for classification on the full and the clean sets (C-Full and C-Clean) and fine-tuned for ranking on the clean set (R-Clean). All these results use the initial regular grid with no RPN.}
\centering
\begin{tabularx}{\textwidth}{@{}p{1.5cm}p{3cm}YYYYY@{}}
\toprule
&&\multicolumn{2}{c}{\bfseries R-MAC}
&\multicolumn{3}{c}{\bfseries Learned R-MAC} \\
\cmidrule(r){3-4} \cmidrule{5-7} 
Dataset & PCA & \cite{Tolias2016} & { Reimp.} & { C-Full} &  { C-Clean} & { R-Clean}\\
\toprule
\multirow{2}{*}{Oxford 5k} & { PCA Paris} & 66.9 & 66.9  & ~~- & ~~- & ~~-\\
&  { PCA Landmarks} &  ~~- & 66.2 & 74.8  & 75.2  & 81.1 \\ 
\midrule
\multirow{2}{*}{Paris 6k} & { PCA Oxford} & 83.0 & 83.0  & ~~- & ~~- & ~~-\\
&  {PCA Landmarks} &  ~~- & 82.3 & 82.5 & 83.2  & 86.0 \\ 
\bottomrule
\end{tabularx}
\label{tab:ml}
\end{table}

\subsection{Datasets and experimental details}
\label{sec:dataset}

\myparagraph{Datasets}
We evaluate our approach on five standard datasets.
We experiment mostly with the \textbf{Oxford 5k} building dataset \cite{Philbin2007} and the \textbf{Paris 6k} dataset
\cite{Philbin2008}, that contain respectively $5,062$ and $6,412$ images. For both datasets there are $55$ query images,
each annotated with a region of interest.
To test instance-level retrieval on a larger-scale scenario, we also consider the \textbf{Oxford 105k} and the
\textbf{Paris 106k} datasets that extend Oxford 5k and Paris 6k with 100k distractor images from
\cite{Philbin2007}.
Finally, the INRIA \textbf{Holidays} dataset~\cite{Jegou2008} is composed of 1,491 images and 500 different scene
queries.

\myparagraph{Evaluation}
For all datasets we use the standard evaluation protocols and report mean Average Precision (mAP).
As is standard practice, in Oxford and Paris one uses only the annotated region of interest of the query, while for Holidays one uses the whole query image.
Furthermore, the query image is removed from the dataset when evaluating on Holidays, but not on Oxford or Paris.

\myparagraph{Experimental details}
Our experiments use the very deep network (VGG16) of Simonyan \etal \cite{Simonyan2014} pre-trained on the ImageNet ILSVRC challenge as a starting point.
All further learning is performed on the Landmarks dataset unless explicitly noted.
To perform fine-tuning with classification \cite{Babenko2014} we follow standard practice and resize the images to
multiple scales (shortest side in the $\left[256-512\right]$ range) and extract random crops of $224\times 224$ pixels.
This fine-tuning process took approximately 5 days on a single Nvidia K40 GPU.
When performing fine-tuning with the ranking loss, it is crucial to mine hard triplets in an efficient manner, as random triplets will mostly produce easy triplets or triplets with no loss. 
As a simple yet effective approach, we first perform a forward pass on approximately ten thousand images to obtain their
representations. We then compute the losses of all the triplets involving those features (with margin $m=0.1$), which is
fast once the representations have been computed. 
We finally sample triplets with a large loss, which can be seen as hard negatives. We use them to train the network with SGD with momentum, with a learning rate of $10^{-3}$ and weight decay of $5\cdot 10^{-5}$. 
Furthermore, as images are large, we can not feed more than one triplet in memory at a time.
To perform batched SGD we accumulate the gradients of the backward passes and only update the weights every $n$ passes, with $n=64$ in our experiments. 
To increase efficiency, we only mine new hard triplets every $16$ network updates. 
Following this process, we could process approximately $650$ batches of $64$ triplets per day on a single K40 GPU.
We processed approximately $2000$ batches in total, \ie, 3 days of training.
To learn the RPN, we train the net for $200$k iterations with a weight decay of $5\cdot 10^{-5}$ and a learning rate of $10^{-3}$, which is decreased by a factor of $10$ after $100$k iterations.
This process took less than $24$ hours.

\subsection{Influence of fine-tuning the representation}
\label{sec:exp-ml}

In this section we report retrieval experiments for the baselines and our ranking loss-based approach.
All results are summarized in Table \ref{tab:ml}.
First of all, as can be seen in the first and second columns, the accuracy of our reimplementation of R-MAC is
identical to the one of the original paper.  We would also like to highlight the following points:

\myparagraph{PCA learning} R-MAC \cite{Tolias2016} learns the PCA on different datasets depending on the target dataset
(\ie learned on Paris when evaluating on Oxford and vice versa).
A drawback of this is that different models need to be generated depending on the target dataset.
Instead, we use the Landmarks dataset to learn the PCA. This leads to a slight decrease in performance, but allows us to have a single universal model that can be used for all datasets.

\myparagraph{Fine-tuning for classification} We evaluate the approach of Babenko \etal \cite{Babenko2014}, where the
original network pre-trained on ImageNet is fine-tuned on the Landmarks dataset on a classification task.  We fine-tune
the network with both the complete and the clean versions of Landmarks, denoted by \emph{C-Full} and \emph{C-Clean} in the table.
This fine-tuning already brings large improvements over the original results.
Also worth noticing is that, in this case, cleaning the dataset seems to bring only marginal improvements over using the complete dataset. 

\begin{figure}[t!]
\resizebox{\linewidth}{!}{%
\begin{tabular}{ccc}
\includegraphics[bb=00bp 0bp 320bp 288bp,clip,height=0.3\linewidth]{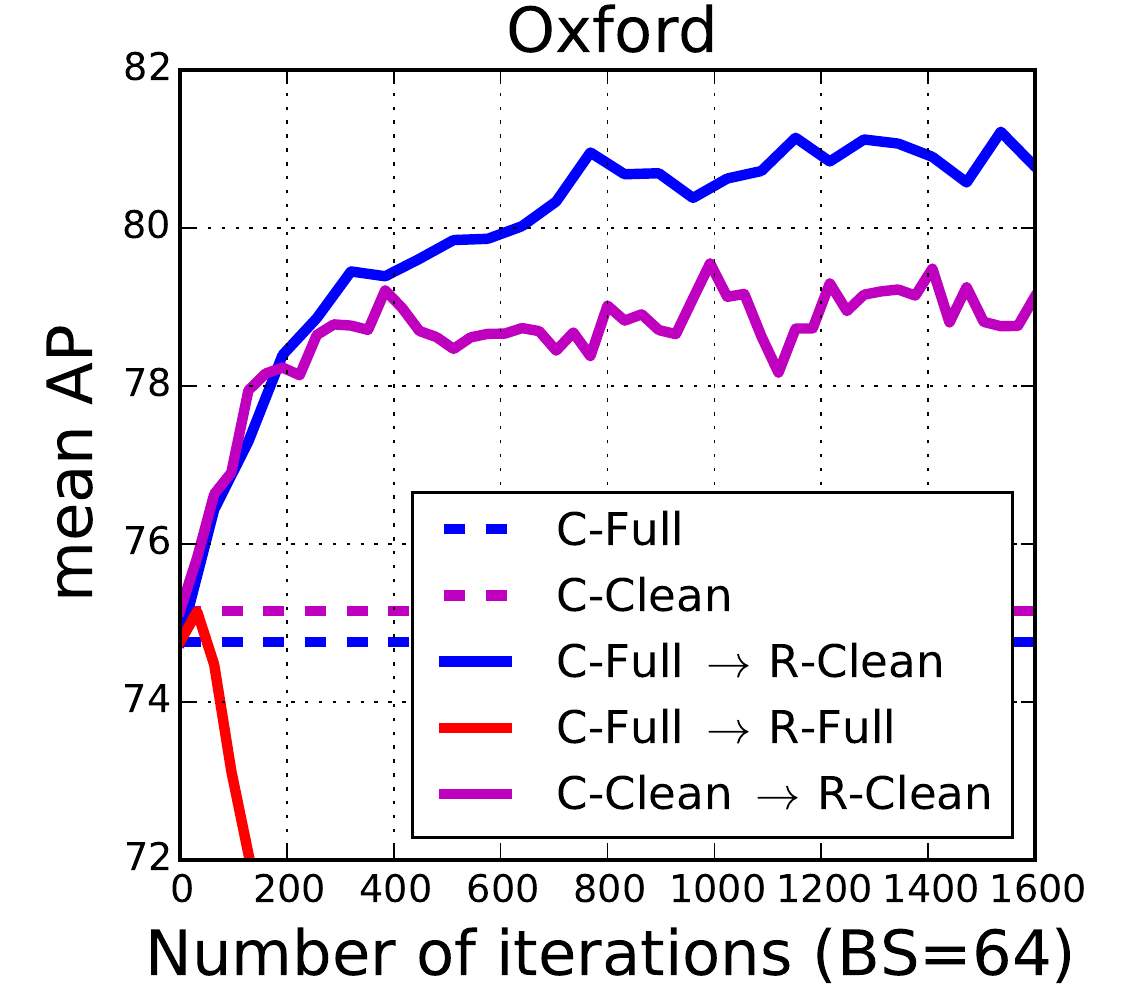} & \includegraphics[bb=00bp 0bp 340bp 288bp,clip,height=0.3\linewidth]{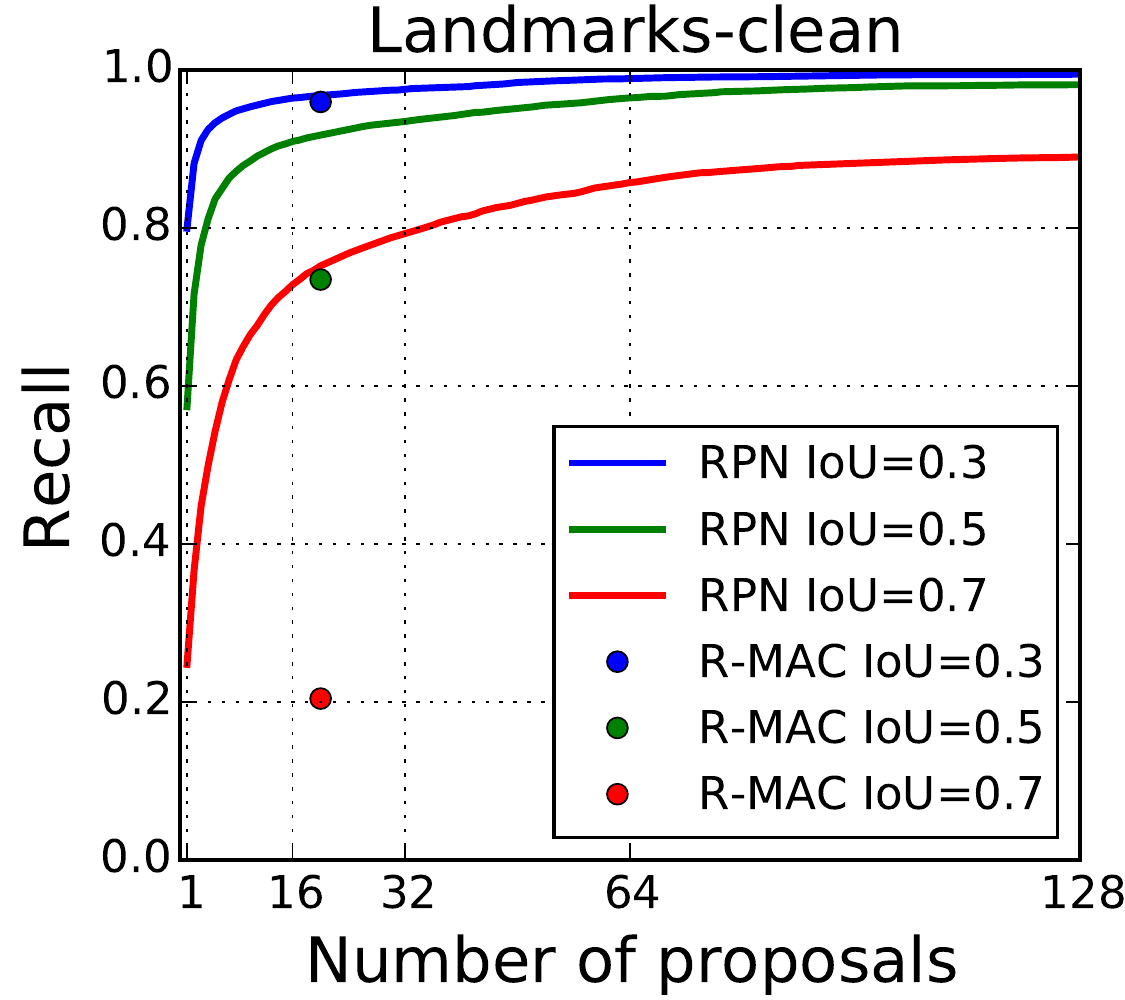} & 
\resizebox{!}{0.28\linewidth}{\includegraphics{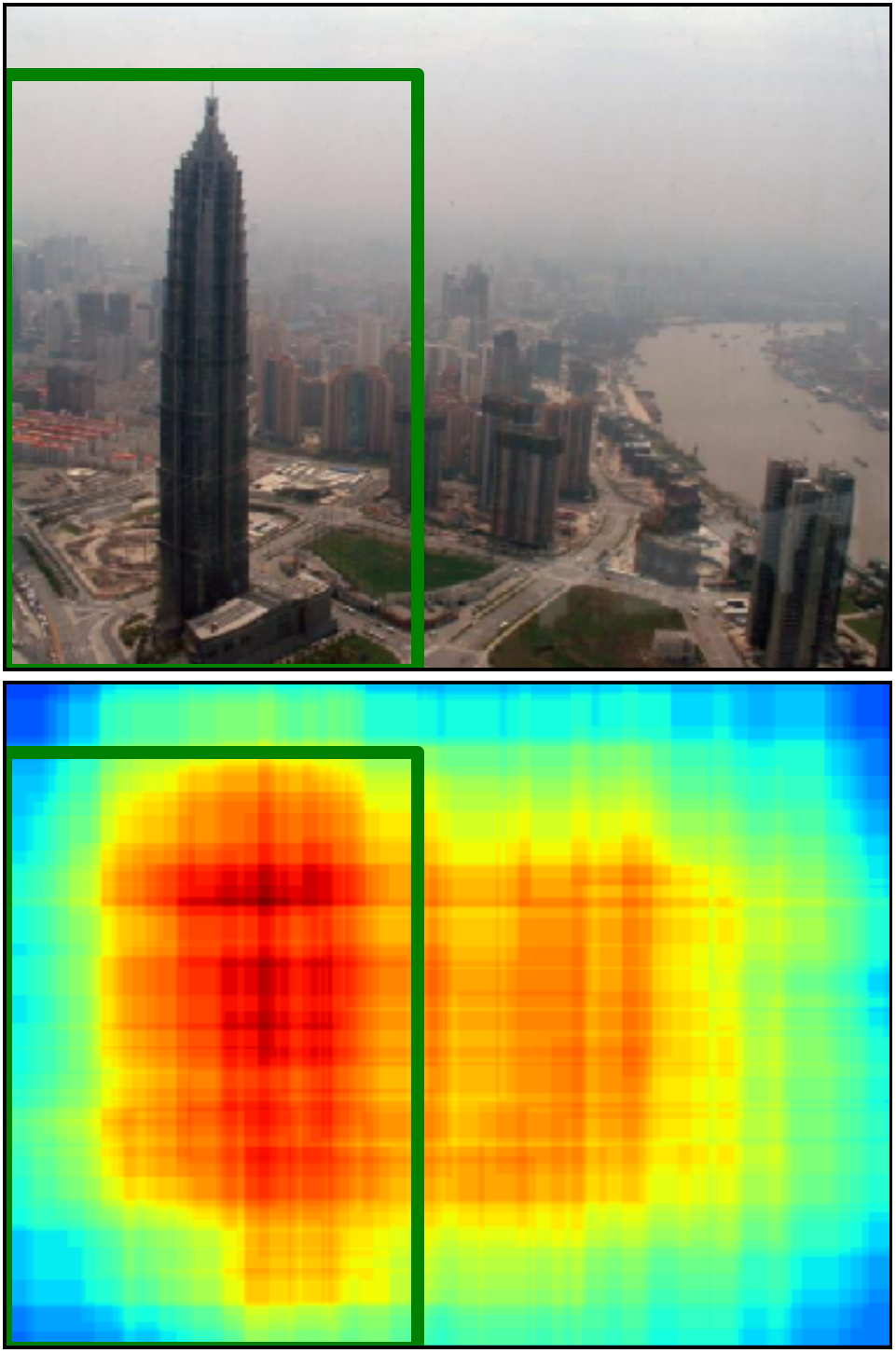}\includegraphics{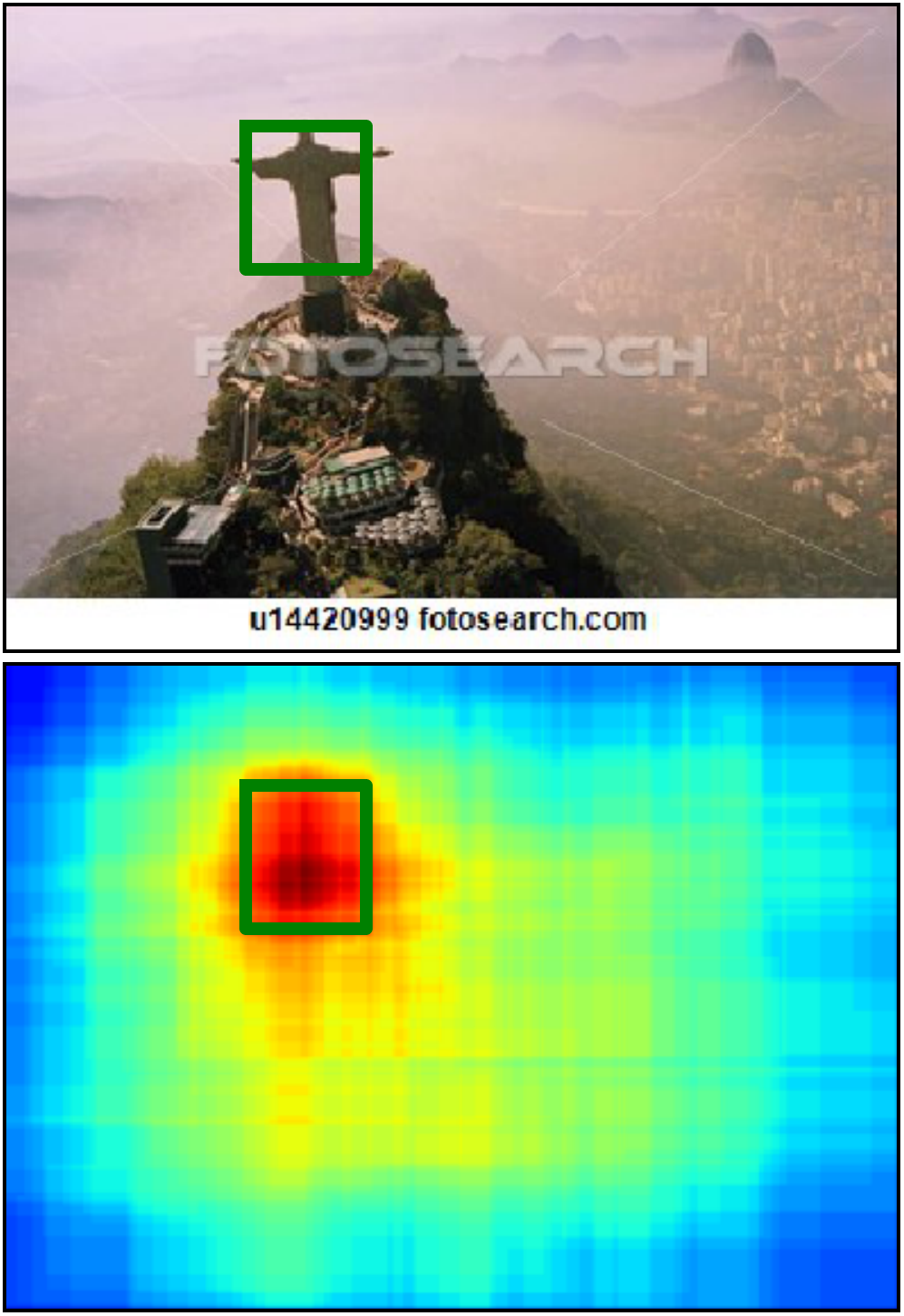}\includegraphics{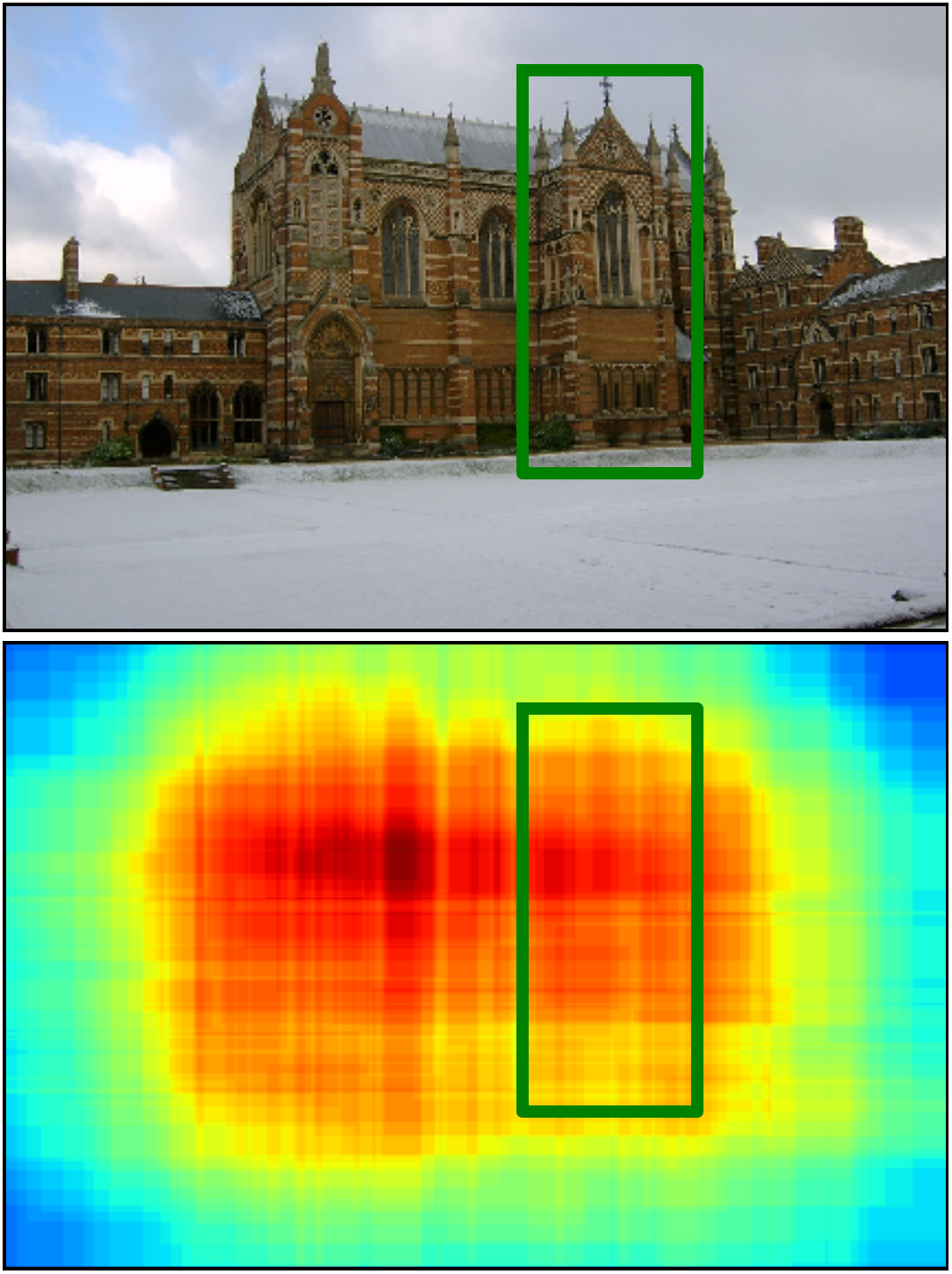}\includegraphics{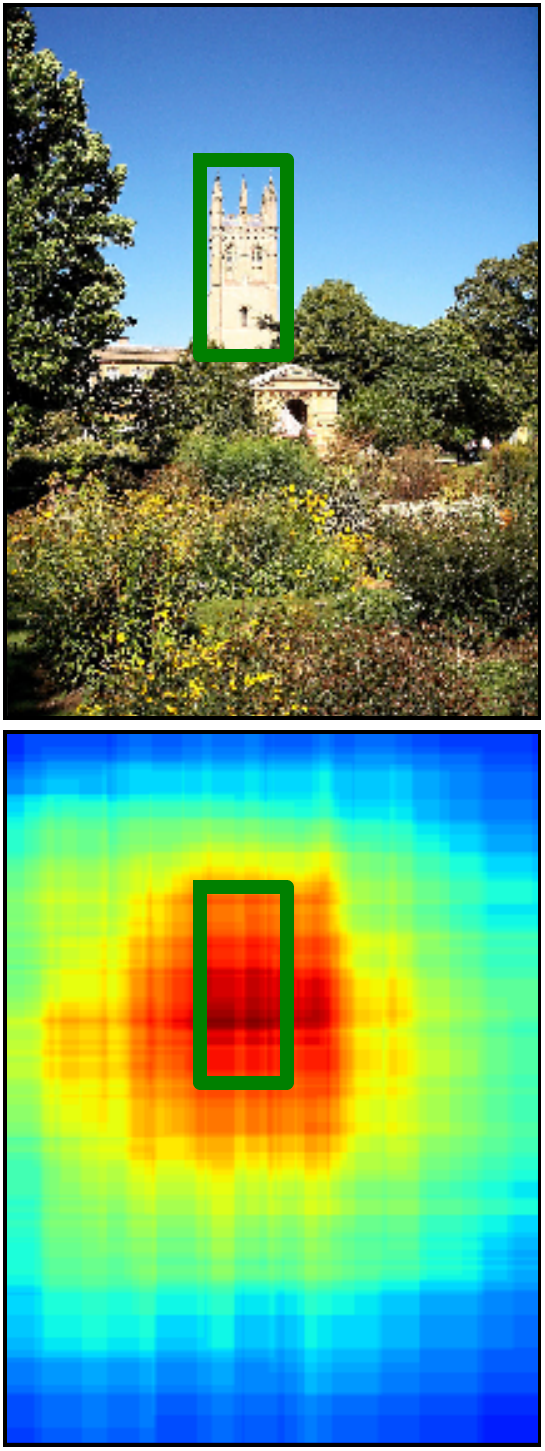}}
\tabularnewline
\tabularnewline
\end{tabular}
}
\caption{\textbf{Left}: evolution of mAP when learning with a rank-loss for different initializations and training sets. 
\textbf{Middle}: landmark detection recall of our learned RPN for several IoU thresholds 
compared to the R-MAC fixed grid.
 \textbf{Right}: heat-map of the coverage achieved by our proposals on images from the
 Landmark and the Oxford 5k datasets. 
 Green rectangles are ground-truth bounding boxes. }
 \label{fig:plots_misc_and_props}
 \end{figure}

  \myparagraph{Fine-tuning for retrieval} We report results using the proposed ranking loss
 (Section~\ref{sec:learning}) in the last column, denoted by \emph{R-Clean}.  We observe how this brings consistent
 improvements over using the less-principled classification fine-tuning.
 Contrary to the latter, we found of paramount
 importance to train our Siamese network using the clean dataset, as the triplet-based training process is less
 tolerant to outliers. Fig. \ref{fig:plots_misc_and_props} (left) illustrates these findings by plotting
 the mAP obtained on Oxford 5k at several training epochs for different settings.
 It also shows the importance of initializing the network with a model that was first
 fine-tuned for classification on the full landmarks dataset. Even if \emph{C-Full} and \emph{C-Clean} 
 obtain very similar scores, we speculate that the model trained with the full Landmark dataset has seen more diverse images so its weights are a better starting point. 

 \myparagraph{Image size} R-MAC \cite{Tolias2016} finds important to use high resolution images (longest side resized to $1024$ pixels).
 In our case, after fine-tuning, we found no noticeable difference in accuracy between $1024$ and $724$ pixels.
 All further experiments resize images to $724$ pixels, significantly speeding up the image encoding and training.

 \begin{table}[t!]
 \footnotesize
 \caption{\textbf{Proposals network.} mAP results for Oxford 5k and Paris 6k obtained with a fixed-grid R-MAC,
 and our proposal network, for an increasingly large number of proposals, before and after fine-tuning with a ranking-loss. The rigid grid extracts, on average, 20 regions per image.}
 \centering
 \begin{tabularx}{\textwidth}{@{}p{2cm}p{2cm}p{1.4cm}YYYYYY@{}}
 \toprule
 & & & \multicolumn{6}{c}{\bfseries \# Region Proposals} \\
 \cmidrule{4-9} 
 Dataset & Model & {\bfseries Grid} & 16 &  32 & 64 & 128 & 192 & 256\\
 \midrule
 \multirow{2}{*}{Oxford 5k} & C-Full & 74.8 & 74.9 & 75.3 & 75.3 & 76.4 & 76.7 & 76.8 \\
 & R-Clean & 81.1 & 81.5 &  82.1 & 82.6 & 82.8 & 83.1 & 83.1 \\ 
 \midrule
 \multirow{2}{*}{Paris 6k} & C-Full & 82.5 & 81.8 & 82.8 & 83.4 & 84.0 & 84.1 & 84.2\\
 & R-Clean & 86.0 & 85.4 & 86.2 & 86.7 & 86.9 & 87.0 & 87.1\\ 
 \bottomrule
 \end{tabularx}
 \label{tab:proposals}
 \end{table}

 \subsection{Evaluation of the proposal network}
 \label{sec:exp-proposal}
 In this section we evaluate the effect of replacing the rigid grid of R-MAC with the regions produced by the proposal network.

 \myparagraph{Evaluating proposals}
 We first evaluate the relevance of the regions predicted by our proposal network.
 Fig.~\ref{fig:plots_misc_and_props} (middle) shows the detection recall obtained in the validation set of Landmarks-Clean for different IoU (intersection over union) levels as a function of the number of proposals, and compares it with the recall obtained by the rigid grid of R-MAC. 
 The proposals obtain significantly higher recall than the rigid grid even when their number is small. 
 This is consistent with the quantitative results (Table \ref{tab:proposals}), where 32-64 proposals already outperform the rigid regions.
 Fig.~\ref{fig:plots_misc_and_props} (right) visualizes the proposal locations as a heat-map on a few sample images of Landmarks and Oxford 5k. It clearly shows that the proposals are centered around the objects of interest.
 For the Oxford 5k images, the query boxes are somewhat arbitrarily defined. In this case, as expected, our proposals naturally align with the entire landmark in a query agnostic way. 

 \myparagraph{Retrieval results}
 We now evaluate the proposals in term of retrieval performance, see Table \ref{tab:proposals}.
 The use of proposals improves over using a rigid grid, even with a baseline model only fine-tuned for classification (\ie without ranking loss).
 On Oxford 5k, the improvements brought by the ranking loss and by the proposals are complementary, increasing the accuracy from $74.8$ mAP with the C-Full model and a rigid grid up to $83.1$ mAP with ranking loss and 256 proposals per image.

 \begin{table}[t!]
 \caption{Accuracy comparison with the  state of the art. Methods marked with an * use the full image as a query in Oxford and Paris instead of using the annotated region of interest as is standard practice. Methods with a $\triangleright$ manually rotate Holidays images to fix their orientation. $^\dagger$ denotes our reimplementation.
 We do not report QE results on Holidays as it is not a standard practice.\label{tab:soa}}
 \footnotesize
 \centering
 \begin{tabularx}{\textwidth}{@{}p{0.5cm}p{3.5cm}p{1cm}YYYYY@{}} \toprule
 & &  & \multicolumn{5}{c}{\bfseries Datasets} \\
 \cmidrule{4-8} 
 & \bfseries{Method} & {\bfseries Dim.}  & Oxf5k & Par6k & Oxf105k & Par106k & Holidays \\
 \midrule 
 \rtext{Global descriptors}{14}&{\scriptsize J\'egou \& Zisserman \cite{jegou:2014} }   &{\scriptsize  1024} & 56.0  &-    &  50.2    & -  & 72.0\\
 &{\scriptsize J\'egou \& Zisserman \cite{jegou:2014} }   & {\scriptsize 128}  & 43.3  & -    &  35.3    & -        & 61.7  \\
 &{\scriptsize Gordo \etal \cite{Gordo2012}}   & {\scriptsize 512}  & -  & -  &  -   & -        & 79.0 \\
 &{\scriptsize Babenko \etal \cite{Babenko2014}} & {\scriptsize 128}  & 55.7*  & -    &  52.3*  & -        & 75.9/78.9$^{\triangleright}$  \\
 &{\scriptsize Gong \etal \cite{Gong2014}}   & {\scriptsize 2048}  & -  & -  &  -   & - & 80.8 \\
 &{\scriptsize Babenko \& Lempitsky\cite{Babenko2015}}   & {\scriptsize 256}  & 53.1  & -    &  50.1    & -        & 80.2$^{\triangleright}$ \\
 &{\scriptsize Ng \etal \cite{Ng2015}}   & {\scriptsize 128}  & 59.3*  & 59.0*  &  -   & -        & 83.6 \\
 &{\scriptsize Paulin \etal \cite{Paulin2015}}   & {\scriptsize 256K}  & 56.5  & -  &  -   & -        & 79.3 \\
 &{\scriptsize Perronnin \& Larlus \cite{Perronnin2015}}   & {\scriptsize 4000}  & -  & -  &  -   & -        & 84.7 \\
 &{\scriptsize Tolias \etal \cite{Tolias2016}}  & {\scriptsize 512}  &66.9& 83.0 &61.6 & 75.7 & 85.2$^{\dagger}$/86.9$^{\dagger,\triangleright}$\\
 &{\scriptsize Kalantidis \etal \cite{Kalantidis2015}} & {\scriptsize 512}& 68.2  & 79.7 & 63.3  & 71.0 & 84.9 \\
 &{\scriptsize Arandjelovic \etal \cite{Arandjelovic2016}} & {\scriptsize 4096}& 71.6 & 79.7 & - & - & 83.1/87.5$^{\triangleright}$  \\
 \cmidrule{2-8} 
 & Previous state of the art & & 71.6 \cite{Arandjelovic2016} & 83.0 \cite{Tolias2016} & 63.3 \cite{Kalantidis2015} & 75.7 \cite{Tolias2016} & 84.9 \cite{Kalantidis2015}\\
 \cmidrule{2-8} 
 &\textbf{Ours} & {\scriptsize 512}  & \textbf{83.1} & \textbf{87.1} & \textbf{78.6} & \textbf{79.7} & \textbf{86.7/89.1$^{\triangleright}$}\\ 
 \midrule
 \midrule
 \rtext{Matching / Spatial verif. / QE}{15}&{\scriptsize Chum \etal \cite{Chum2011}} &&       82.7   & 80.5  &     76.7     &    71.0 & -  \\
 &{\scriptsize Danfeng \etal \cite{Danfeng2011}}&&    81.4   &     80.3  &76.7    &    - 	 &  -    \\
 &{\scriptsize Mikulik \etal \cite{Mikulik2013}} && 84.9 &    82.4  & 79.5 &    77.3   & 75.8$^{\triangleright}$  \\
 &{\scriptsize Shen \etal \cite{Shen2014}} &&        75.2  &      74.1  &     72.9     &     -     & 76.2 \\
 &{\scriptsize Tao \etal \cite{Tao2014}} &&        77.8  &       -    &       -      &       -      & 78.7 \\
 &{\scriptsize Deng \etal \cite{Deng2013}}   &&  84.3  & 83.4  &  80.2   & -        & 84.7 \\
 &{\scriptsize Tolias \etal \cite{Tolias2015}}  & &  86.9  &     85.1  &85.3     &   - &   81.3	 \\
 &{\scriptsize Tolias \etal \cite{Tolias2016}} & {\scriptsize 512} & 77.3  &86.5 &  73.2    &79.8 & -  \\
 &{\scriptsize Tolias \& J\'egou \cite{Tolias2015b}} &&\textbf{89.4} & 82.8 & 84.0  & - & - \\
 &{\scriptsize Xinchao \etal \cite{Xinchao2015}} &&73.7  & - & -  & - & 89.2 \\
 &{\scriptsize Kalantidis \etal \cite{Kalantidis2015}} & {\scriptsize 512}& 72.2  & 85.5 & 67.8  & 79.7 &  - \\
 &{\scriptsize Azizpour \etal \cite{Azizpour2015}} &  & 79.0 & 85.1 & -  & -  & \textbf{90.0} \\
 \cmidrule{2-8} 
 & Previous state of the art & & \textbf{89.4} \cite{Tolias2015b} & 86.5 \cite{Tolias2016} & 85.3 \cite{Tolias2015} & 79.8 \cite{Tolias2016} & \textbf{90.0} \cite{Azizpour2015}\\
 \cmidrule{2-8} 
 &\textbf{Ours + QE} & {\scriptsize 512} & 89.1 & \textbf{91.2} & \textbf{87.3} & \textbf{86.8} & - \\ 
 \bottomrule
 \end{tabularx}
 \end{table}

 \subsection{Comparison with the state of the art}
 \label{sec:exp-final}
 Finally we compare our results with the current state of the art in Table \ref{tab:soa}. 
 In the first part of the table we compare our approach with other methods that also compute global image representations without performing any form of spatial verification or query expansion at test time.
 These are the closest methods to ours, yet our approach significantly outperforms them on all datasets -- in
one case by more than 15 mAP points. This demonstrates that a good underlying representation is important, but also that using features learned for the particular task is crucial.

 In the second part of Table \ref{tab:soa} we compare our approach with other methods that do not necessarily rely on a
 global representation. Many of these methods have larger memory footprints (\eg~\cite{Tolias2015b,Tolias2016,Danfeng2011,Azizpour2015}) and perform a costly spatial verification
 (SV) at test time (\eg~\cite{Tolias2016,Xinchao2015,Mikulik2013}). Most of them also perform query expansion (QE), which is a comparatively cheap strategy that
 significantly increases the final accuracy. We also experiment with average QE~\cite{Chum2007},
 which has a negligible cost (we use the 10 first returned results),
 and show that, despite not requiring a costly spatial verification stage at test time, our method is on
 equal foot or even improves the state of the art on most datasets. 
 The only methods above us are the ones of Tolias and J\'egou~\cite{Tolias2015b} (Oxford 5k) and Azizpour \etal~\cite{Azizpour2015} (Holidays). However, they are both hardly scalable as they require a lot of memory storage and a costly verification (\cite{Tolias2015b} requires a slow spatial verification that takes more than 1s per query, excluding the descriptor extraction time). Without spatial verification, the approach of Tolias and J\'egou~\cite{Tolias2015b} achieves 84.8 mAP in 200ms. In comparison, our approach reaches 89.1 mAP on Oxford 5k for a runtime of 1ms per query and 2kB data per image.
 Other methods such as~\cite{Danfeng2011,Shen2014,Tolias2015} are scalable and obtain good results, but perform some learning on the target dataset, while in our case we use a single universal model.

 \section{Conclusions}
 \label{sec:conclusions}
 We have presented an effective and scalable method for image retrieval that encodes images into compact global signatures that can be compared with the dot-product. 
 The proposed approach hinges upon two main contributions. First, and in contrast to previous works~\cite{Razavian2014,Perronnin2015,Gong2014}, we \emph{deeply}
 train our network for the specific task of image retrieval.
 Second, we demonstrate the benefit of predicting and pooling the likely locations of regions of interest when encoding the images.
 The first idea is carried out in a Siamese architecture~\cite{Hadsell2006} trained with a ranking loss while the second one relies on the successful architecture of region proposal networks~\cite{Ren2015faster}.
 Our approach very significantly outperforms the state of the art in terms of retrieval performance when using global signatures, and is on par or outperforms more complex methods while avoiding the need to resort to complex pre- or post-processing.

 \bibliographystyle{splncs}
 \bibliography{egbib}

\clearpage
\appendix
\section{Qualitative results}\label{s:qualitative}
\begin{figure}[b!]
 \begin{centering}
\includegraphics[width=1\linewidth]{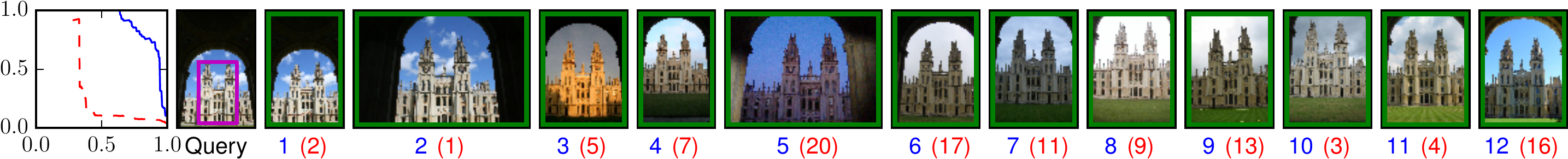}
\includegraphics[width=1\linewidth]{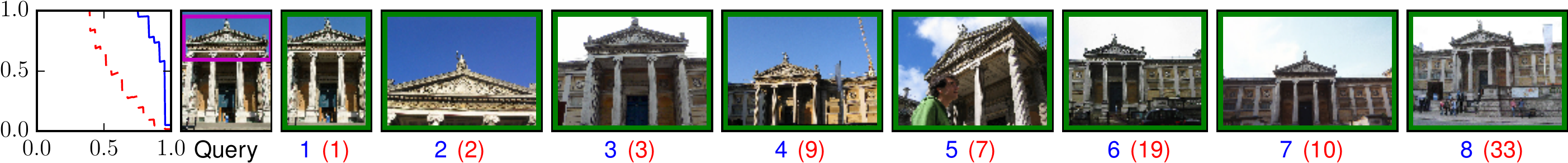}
\includegraphics[width=1\linewidth]{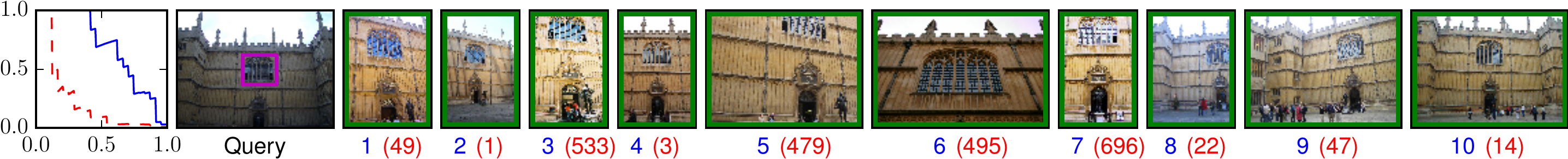}
\includegraphics[width=1\linewidth]{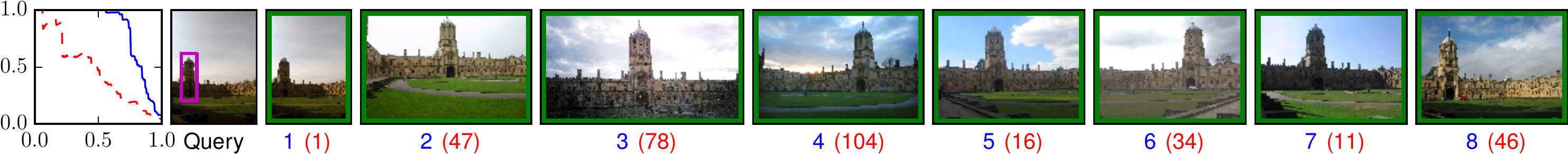}
\includegraphics[width=1\linewidth]{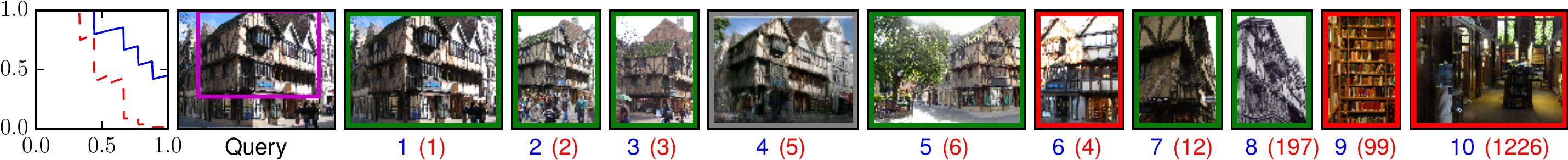}
\includegraphics[width=1\linewidth]{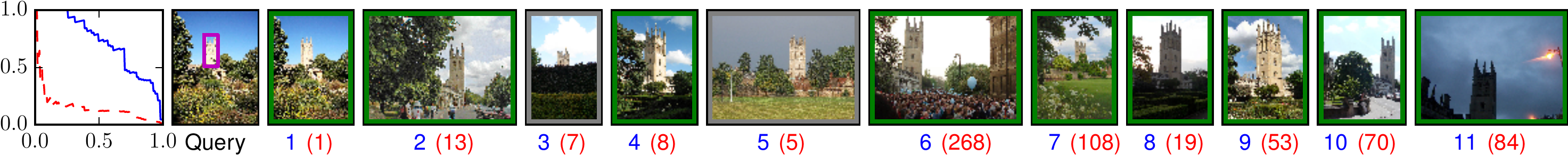}
 \par\end{centering}
 \caption{\label{fig:ox-qual}Top retrieval results and AP curves for a few Oxford queries. R-MAC
 baseline and our method (ranking-loss+proposals) are resp. color-coded as red and blue in the AP plots 
 and in the ranks obtained for each image.
 Green, gray and red borders resp. denote positive, null and negative images.}
 \vspace{-0.3cm}
 \end{figure}

In Figure \ref{fig:ox-qual} we show the top retrieved results by our method, together with AP curves, for a few Oxford 5k queries, and compare them to the results of the R-MAC baseline with VGG16 and no extra training \cite{Tolias2016}. 
The results obtained with the proposed trained model are consistently better in terms of accuracy.
In many cases, several of the correctly retrieved images by our method were not well scored by the baseline method, that placed them far down in the list of results. 
Note also the bad annotation of one of the images in the fifth query (Corn Market), incorrectly labeled as not relevant.

\begin{figure}[t!]
\centering
 \includegraphics[width=1\linewidth]{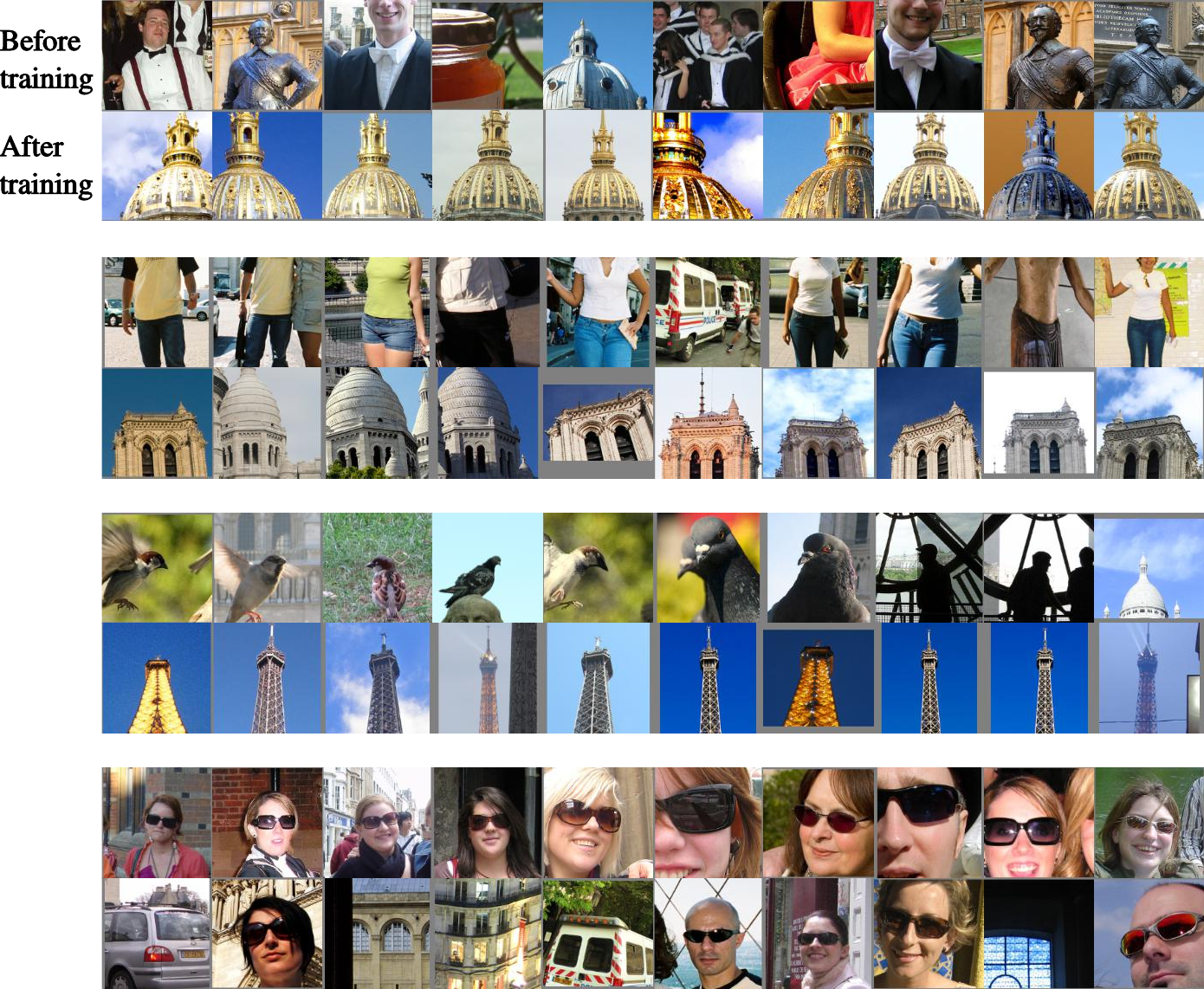}
 \caption{\label{fig:activations} Image patches with largest activation values for some neurons of layer ``conv5\_3'' from VGG16, before (top) and after (bottom) training.}
 \end{figure}

In Figure \ref{fig:activations} we show the image patches that produce the largest activations for several neurons of VGG16's ``conv5\_3'' layer, before and after the proposed training.
First we can observe that, before training, many neurons tend to activate on ``semantic'' patches such as shoulders / bow ties, waists, or sunglasses, even when they do not belong to the same instance, which is not desirable for the task of instance-level retrieval.
After training, many of these neurons have been repurposed to a different task, \eg, shoulders becoming domes.
Many of the new activations do belong to the same instance, which is more useful for the task of instance retrieval.
Note also how the ``sunglasses'' neuron was not correctly repurposed, suggesting that improvements during training are still possible.

\section{Comparing architectures: VGG16 vs ResNet-50}
The recently proposed very deep residual networks \cite{He2016} have shown outstanding results in many computer vision tasks, clearly outperforming other recent architectures while not being much more demanding in terms of computation -- in fact, the $50$ layer residual network (ResNet-50) has a lower computational cost than the popular VGG16 while still obtaining better accuracies on most tasks.
In this section we compare the accuracy obtained by the VGG16 and ResNet-50 architectures when using our proposed framework.

\begin{table}[t!]
\caption{Comparison of the direct R-MAC \cite{Tolias2016}, and the learned versions fine-tuned for classification (C-Full) and fine-tuned for ranking (R-Clean) using the VGG16 and ResNet-50 architectures, for two image resolutions ($S$). All these results use the initial regular grid with no RPN.}
\centering

\begin{tabularx}{\textwidth}{@{}Yp{2cm}YYYYYY@{}}
\toprule
&& \multicolumn{2}{c}{\bfseries Oxford 5k} &  \multicolumn{2}{c}{\bfseries Paris 6k} &  \multicolumn{2}{c}{\bfseries Holidays} \\
\cmidrule(r){3-4} \cmidrule(r){5-6}  \cmidrule{7-8} 
&& $S=724$ & $S=1024$ & $S=724$ & $S=1024$  & $S=724$ & $S=1024$ \\
\cmidrule(r){3-3} \cmidrule(r){4-4} \cmidrule(r){5-5} \cmidrule(r){6-6} \cmidrule(r){7-7} \cmidrule{8-8}

\multirow{2}{*}{Direct} & VGG16 & 59.8 & 66.2 & 79.7 & 82.3 & 85.5 & 87.9 \\
 & ResNet-50 & 69.5 & 69.5 & 84.0 & 83.5 & 90.9 & 93.3\\
\midrule
\multirow{2}{*}{C-Full} & VGG16 & 74.8 & 73.4 & 82.5 & 82.7 & 86.6 & 89.3 \\
 & ResNet-50 & 75.4 & 76.1 & 86.1 & 85.5 & 93.2 & 93.4\\
\midrule
\multirow{2}{*}{R-Clean} & VGG16 & 81.1 & - & 86.0 & - & 87.6 & -\\
 & ResNet-50 &  84.5 & - & 90.6 & - & 93.7 & - \\
\midrule
\end{tabularx}
\label{tab:mlresnet}
\end{table}

  \begin{table}[h!]
 \footnotesize
 \caption{\textbf{Proposals network.} Mean AP accuracy on Oxford 5k, Paris 6k, and Holidays, obtained with a fixed-grid R-MAC or with our proposal network, for an increasingly large number of proposals, using the VGG16 and ResNet-50 architectures. All results are after fine-tuning with a ranking loss.}
 \centering
 \begin{tabularx}{\textwidth}{@{}p{2cm}p{2cm}p{1.4cm}YYYYYY@{}}
 \toprule
 & & & \multicolumn{6}{c}{\bfseries \# Region Proposals} \\
 \cmidrule{4-9} 
 Dataset & Model & {\bfseries Grid} & 16 &  32 & 64 & 128 & 192 & 256\\
 \midrule
 \multirow{2}{*}{Oxford 5k} & VGG16 & 81.1 & 81.5 &  82.1 & 82.6 & 82.8 & 83.1 & 83.1 \\ 
& ResNet-50 & 84.5 & 83.7 & 84.0 & 84.4 & 84.5 & 84.6 & 84.5 \\
 \midrule
 \multirow{2}{*}{Paris 6k} &  VGG16 & 86.0 & 85.4 & 86.2 & 86.7 & 86.9 & 87.0 & 87.1\\ 
& ResNet-50 & 90.6 & 90.3 & 90.9 & 91.0 & 91.3 & 91.3 & 91.2\\ 
 \midrule
 \multirow{2}{*}{Holidays} &  VGG16 & 87.6 & 85.4 &87.1 & 88.6 & 89.0 & 89.1 & 89.1 \\ 
	& ResNet-50 & 93.7 & 91.9 & 92.8 & 93.9 & 94.1 & 94.2 & 94.2 \\ 
\bottomrule
 \end{tabularx}
 \label{tab:proposalsresnet}
 \end{table}

  Table \ref{tab:mlresnet} compares the results obtained with VGG16 and ResNet-50 on three settings: without any specific training, with training aimed at classification (C-full), and with training aimed at retrieval (R-clean). In all cases we use a rigid grid with no proposals.
ResNet-50 has a very noticeable lead, particularly when using the baseline approach. Even at lower resolutions, ResNet-50 performs very well and clearly outperforms VGG16.
After training this gap is reduced, but it is still clear that ResNet-50 obtains a significant advantage with respect to VGG16, despite being faster at testing time.

Table \ref{tab:proposalsresnet} compares VGG16 and ResNet-50 after replacing the rigid grid with proposals obtained with a region proposal network.
Somewhat surprisingly, proposals on the residual network have a smaller impact. 
We believe that there are two reasons for that.
First, ResNet-50 seems to be better at leveraging the rigid grid information  (ResNet-50 with a grid is already better than VGG16 with proposals) and therefore may not need the extra granularity that the proposals provide.
Second, the final activation map of ResNet-50 is half the size of the activation map of VGG16, which leads to less accurate localizations.
To address this last issue one would need to change some aspects of the ResNet-50 architecture, which is out of the scope of this work.
Still, in all cases, proposals always lead to an improvement in accuracy and to better results than VGG16.

Finally, Table \ref{tab:soaresnet} compares the results of the VGG16 and ResNet-50 networks with the current state of the art, including works that appeared after the original ECCV 2016 submission. In Oxford 5k, both VGG16 and ResNet-50 obtain similar results.
However, in all remaining datasets, ResNet-50 obtains significantly better results.
Both VGG16 and ResNet-50 clearly outperform all fixed-length representation methods, even those published after ECCV, and obtains comparable or better results than more complex and costly methods.

\begin{table}[h]
   \renewcommand{\arraystretch}{0.8} 
 \caption{Accuracy comparison with the  state of the art. Methods marked with an * use the full image as a query in Oxford and Paris instead of using the annotated region of interest as is standard practice. Methods with a $\triangleright$ manually rotate Holidays images to fix their orientation. $^\dagger$ denotes our reimplementation.
 We do not report QE results on Holidays as it is not a standard practice.\label{tab:soaresnet}}
 \footnotesize
 \centering
 \begin{tabularx}{\textwidth}{@{}p{0.5cm}p{3.7cm}p{0.8cm}YYYYY@{}} \toprule
 & &  & \multicolumn{5}{c}{\bfseries Datasets} \\
 \cmidrule{4-8} 
 & \bfseries{Method} & {\bfseries Dim.}  & Oxf5k & Par6k & Oxf105k & Par106k & Holidays \\
 \midrule 
 \rtext{Global descriptors}{17}&{\scriptsize J\'egou \& Zisserman \cite{jegou:2014} }   &{\scriptsize  1024} & 56.0  &-    &  50.2    & -  & 72.0\\
 &{\scriptsize J\'egou \& Zisserman \cite{jegou:2014} }   & {\scriptsize 128}  & 43.3  & -    &  35.3    & -        & 61.7  \\
 &{\scriptsize Gordo \etal \cite{Gordo2012}}   & {\scriptsize 512}  & -  & -  &  -   & -        & 79.0 \\
 &{\scriptsize Babenko \etal \cite{Babenko2014}} & {\scriptsize 128}  & 55.7*  & -    &  52.3*  & -        & 75.9/78.9$^{\triangleright}$  \\
 &{\scriptsize Gong \etal \cite{Gong2014}}   & {\scriptsize 2048}  & -  & -  &  -   & - & 80.8 \\
 &{\scriptsize Babenko \& Lempitsky\cite{Babenko2015}}   & {\scriptsize 256}  & 53.1  & -    &  50.1    & -        & 80.2$^{\triangleright}$  \\
 &{\scriptsize Ng \etal \cite{Ng2015}}   & {\scriptsize 128}  & 59.3*  & 59.0*  &  -   & -        & 83.6 \\
 &{\scriptsize Paulin \etal \cite{Paulin2015}}   & {\scriptsize 256K}  & 56.5  & -  &  -   & -        & 79.3 \\
 &{\scriptsize Perronnin \& Larlus \cite{Perronnin2015}}   & {\scriptsize 4000}  & -  & -  &  -   & -        & 84.7 \\
 &{\scriptsize Tolias \etal \cite{Tolias2016}}  & {\scriptsize 512}  &66.9&83.0&61.6 &75.7 & 85.2$^{\dagger}$/86.9$^{\dagger,\triangleright}$\\
 &{\scriptsize Kalantidis \etal \cite{Kalantidis2015}} & {\scriptsize 512}& 68.2  & 79.7 & 63.3  & 71.0 &  84.9 \\
 &{\scriptsize Arandjelovic \etal \cite{Arandjelovic2016}} & {\scriptsize 4096}& 71.6 & 79.7 & - & - & 83.1/87.5$^{\triangleright}$  \\
 &{\scriptsize Radenovic \etal \cite{Radenovic2016}} & {\scriptsize 512}& 79.7 & 83.8 & 73.9 & 76.4 & 82.5$^{\triangleright}$ \\
 \cmidrule{2-8} 
 & Previous state of the art & & 79.7 \cite{Radenovic2016} & 83.8 \cite{Radenovic2016} & 73.9 \cite{Radenovic2016} & 76.4 \cite{Radenovic2016} & 84.9 \cite{Kalantidis2015}\\
 \cmidrule{2-8} 
 &{\footnotesize \textbf{Ours [VGG16]}} & {\scriptsize 512}  & 83.1 & 87.1 & 78.6 & 79.7 & 86.7/89.1$^{\triangleright}$\\ 
 &{\footnotesize \textbf{Ours [ResNet50]}} & {\scriptsize 2048}  & \textbf{84.5} & \textbf{91.2} & \textbf{81.6} & \textbf{86.3} & \textbf{90.7/94.2$^{\triangleright}$}\\ 
 \midrule
 \midrule
 \rtext{Matching / Spatial verif. / QE}{17}&{\scriptsize Chum \etal \cite{Chum2011}} &&       82.7   & 80.5  &     76.7     &    71.0 & -  \\
 &{\scriptsize Danfeng \etal \cite{Danfeng2011}}&&    81.4   &     80.3  &76.7    &    - 	 &  -    \\
 &{\scriptsize Mikulik \etal \cite{Mikulik2013}} && 84.9 &    82.4  & 79.5 &    77.3   & 75.8$^{\triangleright}$   \\
 &{\scriptsize Shen \etal \cite{Shen2014}} &&        75.2  &      74.1  &     72.9     &     -     & 76.2 \\
 &{\scriptsize Tao \etal \cite{Tao2014}} &&        77.8  &       -    &       -      &       -      & 78.7 \\
 &{\scriptsize Deng \etal \cite{Deng2013}}   &&  84.3  & 83.4  &  80.2   & -        & 84.7 \\
 &{\scriptsize Tolias \etal \cite{Tolias2015}}  & &  86.9  &     85.1  &85.3     &   - &   81.3	 \\
 &{\scriptsize Tolias \etal \cite{Tolias2016}} & {\scriptsize 512} & 77.3  &86.5 &  73.2    &79.8 & -  \\
 &{\scriptsize Tolias \& J\'egou \cite{Tolias2015b}} &&\textbf{89.4} & 82.8 & 84.0  & - & - \\
 &{\scriptsize Xinchao \etal \cite{Xinchao2015}} &&73.7  & - & -  & - & 89.2 \\
 &{\scriptsize Kalantidis \etal \cite{Kalantidis2015}} & {\scriptsize 512}& 72.2  & 85.5 & 67.8  & 79.7 &  - \\
 &{\scriptsize Radenovic \etal \cite{Radenovic2016}} & {\scriptsize 512}& 85.0 &86.5 & 81.8 & 78.8  & - \\
 &{\scriptsize Azizpour \etal \cite{Azizpour2015}} &  & 79.0 & 85.1 & -  & -  & \textbf{90.0} \\
 \cmidrule{2-8} 
 & Previous state of the art & & \textbf{89.4} \cite{Tolias2015b} & 86.5 \cite{Tolias2016} & 85.3 \cite{Tolias2015} & 79.8 \cite{Tolias2016} & \textbf{90.0} \cite{Azizpour2015}\\
 \cmidrule{2-8} 
 &{\footnotesize \textbf{Ours + QE [VGG16]}} & {\scriptsize 512} & 89.1 & 91.2 & 87.3 & 86.8 & - \\ 
 &{\footnotesize \textbf{Ours + QE [ResNet50]}} & {\scriptsize 2048} & 89.0  & \textbf{93.8} & \textbf{87.8} & \textbf{90.5} & - \\ 
 \bottomrule
 \end{tabularx}
 \end{table}

 \end{document}